\documentclass[runningheads]{llncs}

\usepackage[paperwidth=152mm,paperheight=235mm,margin=15mm]{geometry}

\usepackage{eccvabbrv}

\usepackage[T1]{fontenc}
\usepackage{amsmath}
\usepackage{amssymb}
\usepackage{cite}
\usepackage[labelfont=bf,font=small,tableposition=bottom]{caption}
\usepackage[skip=3pt]{subcaption}
\usepackage{graphicx}
\usepackage{booktabs}
\usepackage{pgf}
\usepackage{import}
\usepackage{calc}
\usepackage{multirow} 

\usepackage[accsupp]{axessibility}  

\usepackage[breaklinks,colorlinks,citecolor=blue,linkcolor=blue,urlcolor=blue]{hyperref}

\usepackage[capitalize]{cleveref}
\crefname{section}{Sec.}{Secs.}
\Crefname{section}{Section}{Sections}
\crefname{table}{Tab.}{Tabs.}
\Crefname{table}{Table}{Tables}

\usepackage{orcidlink}

\newcommand{\set}[1]{\left\{ #1 \right\}}
\newcommand{\given}{\,\middle|\,}
\newcommand{\norm}[1]{\lVert #1 \rVert}

\begin{document}

\title{CLEAR-NeRF: Collinearity and Local-region Enhanced Accurate 3D Reconstruction in Unbounded Scenes}

\titlerunning{CLEAR-NeRF}

\author{Vladislav Polianskii\orcidlink{0000-0001-9805-0388} \and
Elijs Dima\orcidlink{0000-0002-4967-3033} \and
Isabel Salmerón Marazuela \and\\
Gergő László Nagy\orcidlink{0009-0002-9479-6240} \and
Sigurdur Sverrisson \and
Volodya Grancharov
}

\authorrunning{V.~Polianskii et al.}

\institute{Ericsson Research, Stockholm, Sweden}

\maketitle

\begin{abstract}
  Many real-world 3D reconstruction applications demand photorealism and metric accuracy across unbounded, complex scenes with challenging lighting and imperfect captures that current Neural Radiance Field (NeRF) pipelines only partly satisfy.
  This study adapts NeRF-based 3D reconstruction to multi-region of interest unbounded scenes to improve robustness to lighting and pose variation while enforcing metric accuracy suitable for digital-twin applications.
  Our approach introduces (i) automated local region localization/detection and reconstruction to seamlessly prioritize areas of interest without proliferating submodules, (ii) collinearity-enforcing ray sampling to learn smooth planar and curved surfaces, (iii) depth-localized neighborhood point extraction to suppress surface artifacts, and (iv) geometry-relevant color aggregation to mitigate lighting- and pose-caused variations.
  Results indicate superior performance of the proposed pipeline over the baseline NeRF models and established Structure from Motion (SfM) - Multi-View Stereo (MVS) solutions.
\end{abstract}

\section{Introduction}
\label{sec:introduction}

3D reconstruction is a crucial technology in the realm of computer vision and graphics, 
enabling both the transformation of real-world objects into digital scene models, 
and the generation of novel representations and views of the digitized scenes.
Recently, Neural Radiance Fields (NeRF)~\cite{mildenhall2021nerf} and 3D Gaussian Splatting (3DGS)~\cite{kerbl20233d} have emerged as dominant methods for scene modelling.
The introduction of NeRF revolutionized photorealistic novel-view synthesis by learning a differentially renderable continuous scene representation from multi-view images. 
Follow-up work has improved efficiency and robustness, \eg~Instant-NGP~\cite{muller2022instant}, mip-NeRF~\cite{barron2021mip}, and extended NeRFs to unbounded scenes, \eg~mip-NeRF 360~\cite{barron2022mip} and NeRF++~\cite{zhang2020nerf++}. 
Frameworks such as Nerfstudio and its Nerfacto model~\cite{tancik2023nerfstudio} consolidate these advances in baselines that can be deployed for applications spanning autonomous driving, robotics, medical imaging, etc.~\cite{molaei2023implicit,wang2024nerfs,xiao2025neural}


Real-world scenes have complex lighting and it is seldom possible to capture perfect 2D images in ideal positions. 
Most real-world scenes contain multiples of relevant objects which need to be reconstructed to a high degree of accuracy, often within an unbounded large-scale environment. 
Conversely, not every object within an unbounded scene is of equal relevance to a particular application; most real-world scenes are a combination of relevant sections and background, with varying requirements on reconstruction accuracy. 
For applications involving digital twinning and real-world decision-making, both fidelity and metric accuracy of the 3D reconstruction is of import.
Therefore, NeRF-based solutions intended for such applications have to address these issues in order to be practically viable.

As NeRF was introduced in~\cite{mildenhall2021nerf} in the context of view synthesis, NeRF-based method surveys have predominantly focused on the visual quality of the 2D images rendered from NeRF, e.g. as seen in \cite{liu2025survey,gao2022nerf,liao2025survey}. 
As Giaquinto \etal \cite{giaquinto2026evaluating} point out, NeRF was not initially designed for metric 3D reconstruction and Unmanned Aerial Vehicle (UAV)-based photogrammetry.
For such use-case, combining Structure-from-Motion (SfM) with bundle adjustment \cite{schonberger2016structure} and Multi-View Stereo (MVS) \cite{seitz2006comparison} has been the go-to approach \cite{berra2020advances,pepe2022uav}. 

In this paper we address the specific challenges of using NeRF for precise 3D point cloud reconstruction with UAV image sets as the reconstruction input.
We focus on handling unbounded scenes with multiple degrees of object relevance while reducing the impact of complex lighting. 
We also ensure that 3D reconstructions are metrically accurate.
We propose the following extensions to NeRF 3D reconstruction and geometry extraction: 
\begin{itemize}
    \item a local-region-focused multi-part scene reconstruction approach to account for multiple areas of interest to allow high-fidelity reconstruction without excessive allocation of NeRF submodules across the scene, 
    \item a change in ray sampling during training to enforce collinearity and thus improve learning of smooth surfaces, both planar and non-planar, 
    \item a change in sampling during 3D geometry extraction to form the 3D point from a depth-localized neighbourhood of pixels and thus reduce artifacts on the reconstructed surfaces,
    \item a change to aggregate ray colours from the geometry-relevant sections of the ray, to reduce the impact of lighting condition variance and deviations in camera poses on reconstructed object colorization.
\end{itemize}
We demonstrate the effect of these alterations on a selection of real-world scenes of complex environments scanned with UAVs and ground LiDARs. This addresses the use-case of 3D reconstruction for applications where metric accuracy of objects is necessary.

\section{Related Work}
\label{sec:related}

NeRF-based methods have been compared against SfM-MVS-based methods in the context of 3D reconstruction accuracy \cite{nex2023benchmarking,lewandowski2023application,mate2025comparison,tsukamoto2025comparative}. 
Mat{\'e}-Gonz{\'a}lez \etal \cite{mate2025comparison}, Nex \etal \cite{nex2023benchmarking} and Tsukamoto \etal \cite{tsukamoto2025comparative} conclude 
that the legacy SfM-MVS methods generate higher, more detailed surface models, while acknowledging the potential in NeRF-based approaches.
Within specific type of scenes - telecom towers recorded by UAVs - Lewandowski \etal \cite{lewandowski2023application} show potential for parity between reference data and NeRF reconstructions created from Mega-NeRF \cite{turki2022mega} and Nerfacto \cite{tancik2023nerfstudio}, given optimal conditions, UAV flight distances, and a tightly bounded scene scale.


\textbf{NeRF-Based 3D Reconstruction of Large-Scale Scenes.} 
Handling large-scale and unbounded scenes is a core challenge for NeRFs in UAV-based photogrammetry. 
Most solutions either manipulate scene space \cite{barron2022mip,wan2024constraining,turki2023suds} or partition scenes into multiple NeRFs \cite{liu2025research,tancik2022block,chen2025block,xiangli2022bungeenerf,xu2023grid,xu2024multi}.
Mip-NeRF 360~\cite{barron2022mip} introduced scene contraction that maps points outside a unit volume into a bounded shell enclosing the unit volume; this is adopted by Nerfacto \cite{tancik2023nerfstudio} and a similar scene compression appears in Nerf++~\cite{zhang2020nerf++}.
Wan \etal~\cite{wan2024constraining} constrain elevation and rescale the vertical axis to boost height fidelity.
Turki \etal~\cite{turki2023suds} compartmentalize scene into static, dynamic and far-field radiance fields and, like Mega-NeRF~\cite{turki2022mega}, first tile large scenes into cells with per-cell models. 
Similar multi-part scene tiling appears in \cite{tancik2022block,xu2024multi,zhang2024aerial,liu2025research}, where scene resolution demands dominate over scene scale.
To address linear scaling of sampling and training costs, Xu \etal~\cite{xu2023grid} use multi-resolution feature grids to steer sampling and augment NeRF positional encodings, along with vertical compaction akin to \cite{wan2024constraining}. 
BungeeNeRF~\cite{xiangli2022bungeenerf} instead adopts a multi-level coarse-to-fine approach, with a multi-stage training process and layered radiance field block structure where fine-level Multilayer Perceptron (MLP) blocks refine the higher-level coarse blocks.

\textbf{Geometric Accuracy of NeRF-Based 3D Reconstruction.} 
Llull \etal~\cite{llull2023evaluation} and Tsukamoto \etal~\cite{tsukamoto2025comparative} highlight the common issue of spatial noise and flat surface deformations in NeRF reconstructions.
Numerous NeRF variants aim to improve quality beyond scaling or scene compartmentalization by focusing on depth~\cite{wang2025gpe,yang2023nerfvs,liao2024improving,fang2025depth}, geometric regularization~\cite{cui20243d,li2025pgc,petrovska2024vision}, and photometric improvements~\cite{zhang2024learning,zhuang2024gtr,liu2025research}. 
Wang \etal~\cite{wang2025gpe} use a Gaussian depth loss from sparse SfM points and replace positional encoding with Gaussian encoding for high-frequency details. 
Yang \etal~\cite{yang2023nerfvs} encourage ray termination at pseudo-depth from depth priors, and regularize depth/color distributions.
Liao \etal~\cite{liao2024improving} adopt depth-centered Gaussian color sampling and image embeddings for robustness to diverse scene lighting.
Fang \etal~\cite{fang2025depth} proposes a modified NeRF ray sampling based on depth confidence, to concentrate samples in geometrically complex regions.
Cui \etal~\cite{cui20243d} impose plane-fitting and Manhattan-world constraints for urban/building-rich scenes. 
Li \etal~\cite{li2025pgc} improve ray density estimation by adjacent-ray correlation and local geometric consistency to reduce surface roughness.
Petrovska and Jutzi~\cite{petrovska2024vision} voxellize the density field and apply a 3D density-gradient Canny filter to improve geometry extraction.
Zhang \etal~\cite{zhang2024learning} employ adaptive random Fourier features and photometric loss from adjacent views to enhance multi-view consistency.
Zhuang \etal~\cite{zhuang2024gtr} decouple color and density MLPs, and separately refine the color MLP to reduce texture-loss.
Liu \etal~\cite{liu2025research} add a texture refinement stage via a dense residual network tailored to scene characteristics.

\section{Method}
\label{sec:method}

\begin{figure}[tb]
  \centering
  \scalebox{0.6}{\def\svgwidth{1.5\linewidth}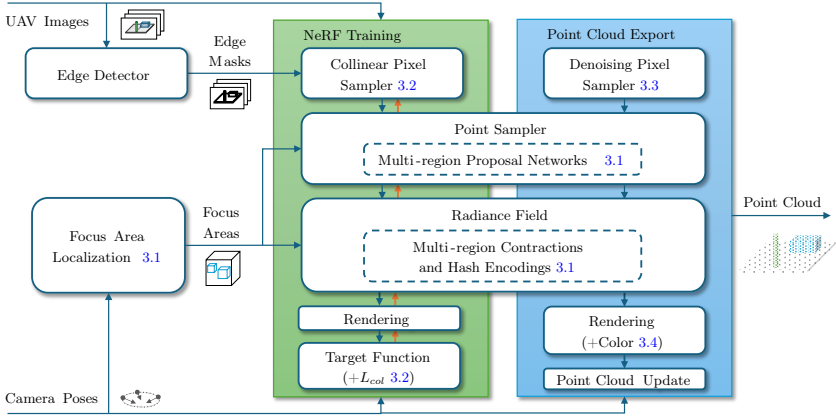}
  \caption{CLEAR-NeRF overview showing NeRF training and point cloud export stages.}
  \label{fig:method_overview}
\end{figure}

Our contribution to precise NeRF-based point cloud reconstruction augments and enhances major parts of the traditional pipeline.
An overview of the method is depicted in Figure~\ref{fig:method_overview}.
Key parts are as follows:  
\begin{itemize}
    \item an automated focus area localization module that detects parts of the scene which require high fidelity reconstruction,
    as well as a modification to the NeRF field. This includes the main field and proposal samplers, allowing for the optimal use of the focus areas (Section~\ref{sec:method_focus_areas}); 
    \item a pixel sampler for the training process together with image preprocessing using edge detector that imposes additional regularization on surface flatness via collinear constraints (Section~\ref{sec:method_collinearity});
    \item an optimized pixel sampler during point cloud extraction. The sampler introduces depth-congruent filtering of incorrectly-sampled points which occur near depth discontinuities (Section~\ref{sec:method_depth});
    \item an alternative color rendering procedure specific for point cloud extraction. The procedure introduces a novel type of integration over a sampled ray to prevent color smearing (Section~\ref{sec:method_color}). 
\end{itemize}

\subsection{Local Region Focused (LRF) Scene Reconstruction}
\label{sec:method_focus_areas}

To address 3D reconstruction of large-scale scenes with sufficient fidelity, solution like \cite{turki2023suds,turki2022mega,tancik2022block,xu2024multi,zhang2024aerial,liu2025research} split the scene into multiple cells as a way to distribute the computational cost and NeRF size.
However, that does not effectively allocate NeRF resources to the relevant areas of the scene, where higher fidelity is required. 
We propose a local region focused scene reconstruction, LRF, consisting of two parts: focus-area localization and multi-resolution NeRF training.

\begin{figure}[tb]
    \centering
    \begin{subfigure}{.32\linewidth}
    \includegraphics[width=\linewidth]{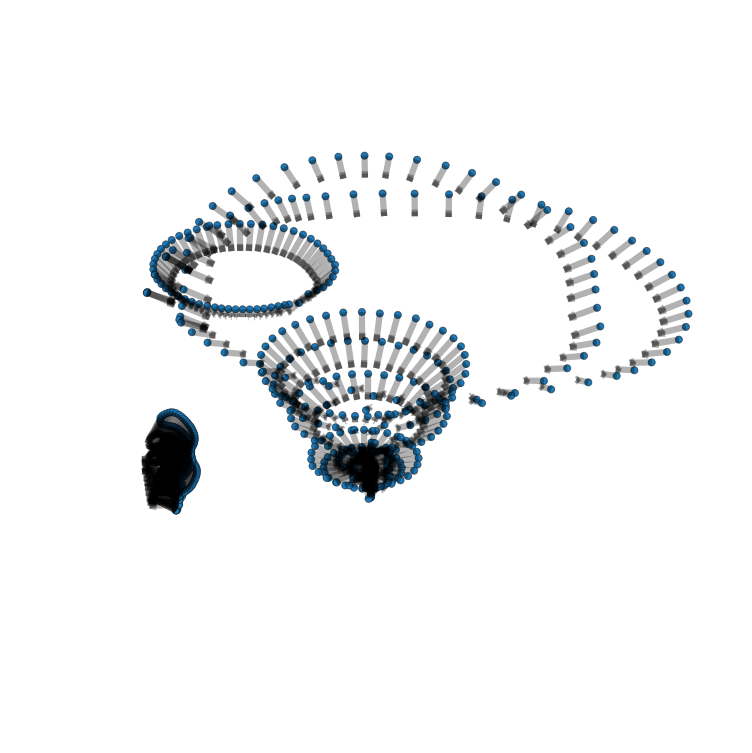}
    \subcaption{Cameras with directions}
    \label{fig:orbit_ex_data}
    \end{subfigure}%
    \hfill
    \begin{subfigure}{.32\linewidth}
    \includegraphics[width=\linewidth]{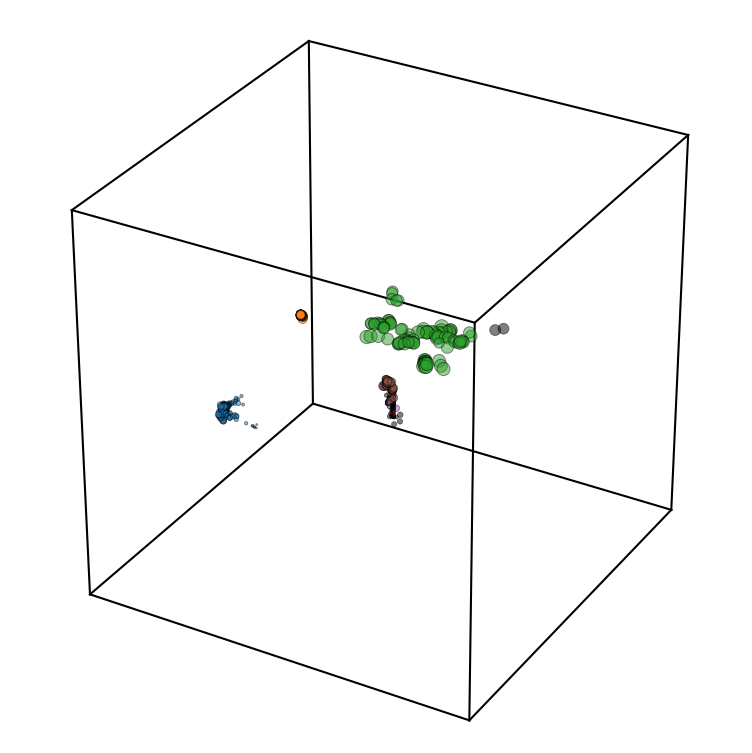}
    \subcaption{Clustering of $\overline{\mathcal{P}}$}
    \label{fig:orbit_ex_clustering}
    \end{subfigure}%
    \hfill
    \begin{subfigure}{.32\linewidth}
    \includegraphics[width=\linewidth]{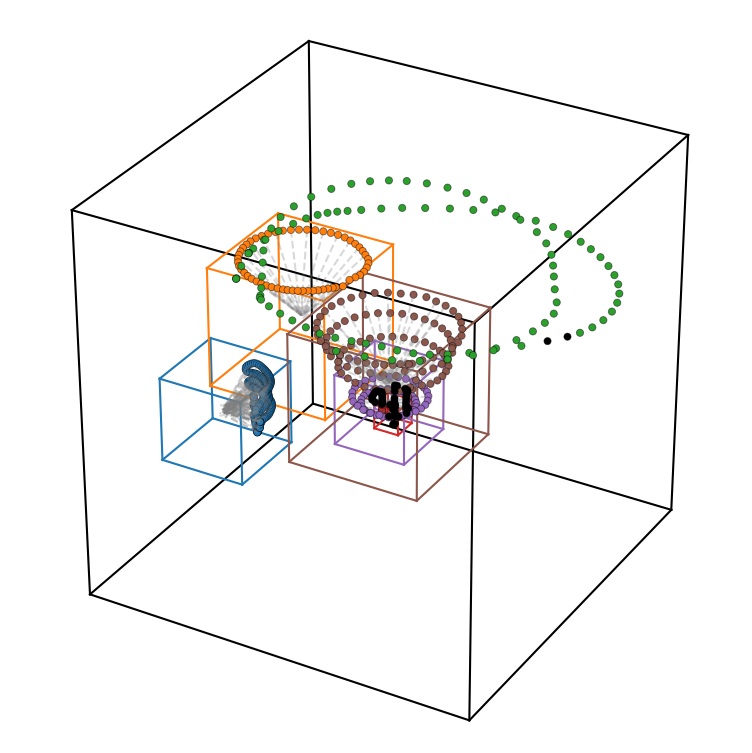}
    \subcaption{Identified focus areas}
    \label{fig:orbit_ex_areas}
    \end{subfigure}
    \caption{An example of automatic focus-area detection for a UAV flight with multiple overlapping and concentric orbits over the scene and objects of interest.}
    \label{fig:orbit_ex}
\end{figure}

\textbf{Focus-area Localization.}
Our way of detecting and localizing focus areas in a large unbounded scene takes inspiration from light caustics, where rays emanate from cameras' optical centers in corresponding directions and we find concentration points along ray bundles, see Figure~\ref{fig:orbit_ex}.

To enable LRF, we first obtain cameras' extrinsic and intrinsic matrices from camera alignment process. Let $\mathcal{K}$ denote the set of all cameras. For each camera $\kappa \in \mathcal{K}$, we have its origin $o_\kappa \in \mathbb{R}^3$ and viewing direction $u_\kappa\in \mathbb{S}^2$, illustrated in Figure~\ref{fig:orbit_ex_data}. 
For a given distance $t \in \mathbb{R}_+$ from camera origin, we define points along the ray from camera $\kappa$ as $x_\kappa(t) = o_\kappa + u_\kappa t$, and the set of all camera origins moving through space along their respective rays as $\mathcal{P}(t) = \set{x_\kappa(t) \given \kappa \in \mathcal{K}}$.

As the first step, for each camera $\kappa$, we find a distance $t_\kappa$ at which the concentration/density for that camera's ray is the highest:
\begin{equation}
t_\kappa = \underset{t \ge 0}{\mathrm{argmin}}{\sum\limits_{x_{\kappa'} (t) \in \text{NN}_{n}(x_\kappa(t)\,|\,\mathcal{P}(t))}\norm{x_\kappa(t) - x_{\kappa'}(t)}_2},
\end{equation}
where $\text{NN}_{n}(x_\kappa(t)\,|\,\mathcal{P}(t))$ are $n$ nearest neighbors of $x_\kappa(t)$ in set $\mathcal{P}(t)$.

Given the distance $t_\kappa$, for every ray we define the 3D point $\hat{x}_\kappa$ indicating where the highest concentration of the other rays occurs as: $\hat{x}_\kappa = x_\kappa(t_\kappa) = o_\kappa + t_\kappa u_\kappa$. 
We also define a 4D set of points $\overline{\mathcal{P}}$ of all such points, to capture how far the cameras are from their respective $\hat{x}_\kappa$, by: $\overline{\mathcal{P}}=\set{\bar{x}_\kappa=(\hat{x}_\kappa;t_\kappa)\given \kappa\in \mathcal{K}}$. 

The next step is to cluster $\overline{\mathcal{P}}$ points to get the focus areas. 
We cluster them using HDBSCAN~\cite{mcinnes2017hdbscan} and the following distance measure between 4D points: 
\begin{equation}
\delta(\bar{x}_\kappa, \bar{x}_{\kappa'})=\max\set{{\lVert \bar{x}_\kappa - \bar{x}_{\kappa'} \rVert}_2 - (t_\kappa + t_{\kappa'})\sin \alpha, 0}.
\end{equation}
The $(t_\kappa + t_{\kappa'})\sin \alpha$ term decreases the distances between points with large offset from the camera; the given semimetric simulates the distance between growing frustums with angular size $\alpha$.
A representation of a 3D projection of $\overline{\mathcal{P}}$ is provided in Figure~\ref{fig:orbit_ex_clustering}, where the resulting clusters are visualized by different colors; the black box outlines the global scene box.

Finally, for each detected cluster of cameras $\mathcal{C}$, the focus area center is identified as $f=\mathbb{E}_{\kappa\in \mathcal{C}}\hat{x}_\kappa$ and the focus area radius as $r=\mathbb{E}_{\kappa\in \mathcal{C}}\norm{f-o_\kappa}_{\infty}$. 
A corresponding cube can be defined as $\prod_{i=1}^{3}[f_i-r, f_i+r]$; this is used as the boundary for local scene contraction.
An example of the resulting focus areas is indicated in Figure~\ref{fig:orbit_ex_areas}.

The proposed process is fully automated and does not require any specific structure of the input image set. The different UAV flight paths do not have to be separate from each other temporally or spatially, instead, the focus areas for LRF 3D reconstruction are inferred from implicit interactions between the estimated camera poses. 



\textbf{Multi-resolution Training Procedure} for global and local scene contractions.
We denote a 3D query position by $p$ and ray direction it is sampled along by $u_{*}$, and we use $M$ to denote the maximum count of the identified focus areas.
We parameterize the radiance field with a global branch ($i=0$) and $0 \le M_0 \le M$ cluster-specific local branches ($i=1,\dots,M_0$).  
Each branch applies its own scene contraction $\tilde{p}=\mathrm{contract}(p)$ defined in~\cite{barron2022mip}
and an independent learnable hash encoding introduced in~\cite{muller2022instant} to produce a feature vector $\mathbf{h}_i(\tilde{p})$. 
We then fuse all inputs by concatenation, $\mathbf{h}(\tilde{p})=\mathrm{concat}(\mathbf{h}_0(\tilde{p}),\dots,\mathbf{h}_{M_0}(\tilde{p}))$, which is fed to the density network to predict $\sigma$ along with an intermediate feature embedding. 
The direction $u_{*}$ is encoded separately with Spherical Harmonics $\mathrm{SH}(u_{*})$ and combined with the density feature in the color network to produce $c$, 
while keeping the encoding parameters of each orbit separate so the model can leverage multiple local views without changing the overall NeRF training procedure.

\begin{figure} [tb]
    \centering
    \includegraphics[width=1\linewidth]{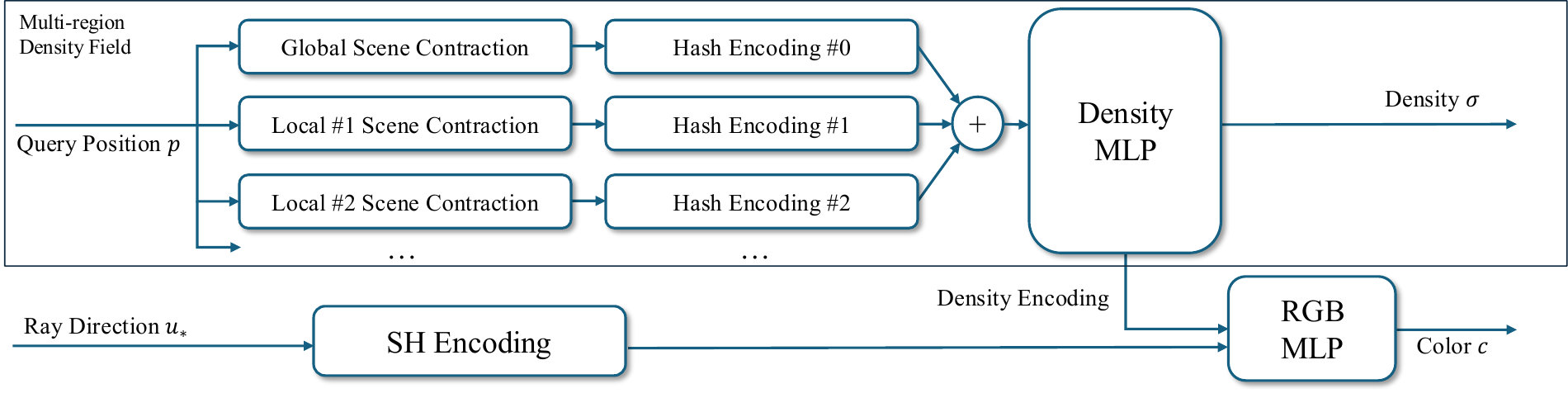}
    \caption{Multi-resolution training procedure overview.}
    \label{fig:multi_res_training}
\end{figure}

By retaining all encodings (\ie global and local) within the concatenated vector $\mathbf{h}$, we ensure that the model learns smooth transitions between the subscenes, as the transition $p \mapsto \mathbf{h}(\tilde{p})$ is entirely continuous. 
An overview of the separation of focus areas during training is shown in Figure~\ref{fig:multi_res_training}.

\subsection{Improved Surface Modeling (ISM) in NeRF-based 3D Scene Digitization}
\label{sec:method_collinearity}
Objects captured from only a few angles, or those with monochromatic textures, can produce deformed surfaces during NeRF reconstruction that visually match the reference photos but fail to represent the correct geometry \cite{tsukamoto2025comparative,llull2023evaluation}. 

Inspired by \cite{cui20243d,li2025pgc}, we propose an improved surface modeling solution, ISM. It is based on an optimization technique by means of an additional loss factor that induces a geometric prior 
onto scene surfaces. This loss penalizes large variations in depth values using a collinearity constraint, and as a result mitigates surface deformations and improves metric 
accuracy of the 3D scene.

\begin{figure} [tb]
    \centering
    \includegraphics[width=.6\linewidth]{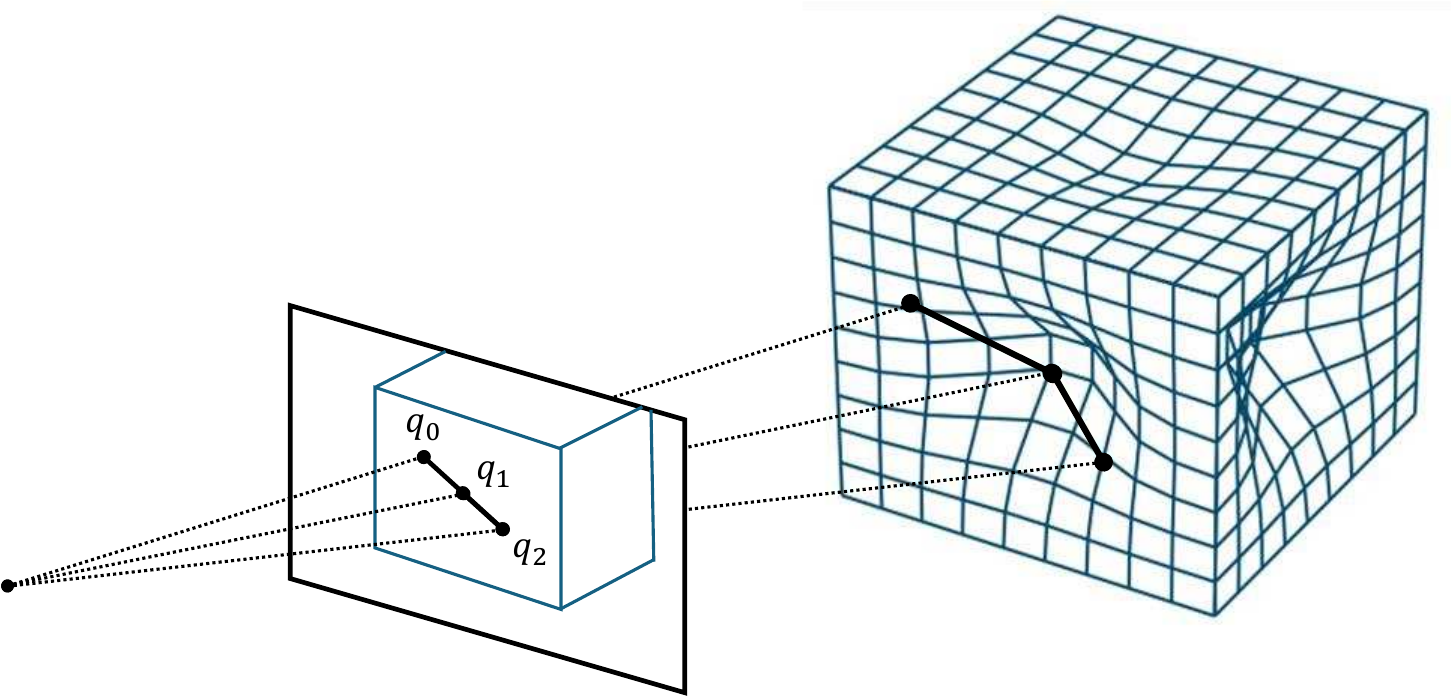}
    \caption{Ray sampling through three selected points on the image plane. Right side shape represents a currently reconstructed 3D scene.}
    \label{fig:ray_sampling_wavy}
\end{figure}

Before the training starts, we use a preprocessing block that computes edges on the input images, such as Canny edge detector~\cite{canny1986computational}.
Then during training, our sampling technique performs random sampling of batches of image pixels in groups of three, like depicted in Figure~\ref{fig:ray_sampling_wavy}. 
Each triplet ($q_0, q_1,q_2$) of pixels denotes a single segment on one image, with $q_1$ being located between $q_0$ and $q_2$. 
The $q_1$ pixel is sampled uniformly first; if it does not land on an edge, the other two pixels $q_0$ and $q_2$ are sampled in such way that $q_1$ is the midpoint between them, the segment $[q_0; q_2]$ does not cross any edges and $\norm{q_0-q_2}$ is a large as possible (below a threshold of 40 pixels). 
If $q_1$ lands on an edge, $q_0$ and $q_2$ are sampled uniformly as well, and the triplet is not considered for the rest of this section. 
By such sampling, we ensure that at least one third of the pixels are sampled uniformly in the image. 


Given the triplet of pixels, we propose a collinearity loss $L_{col}$ to mitigate the reconstruction of deformed surfaces.
A general formula for the value of the loss $L_{col}$ for a triplet of points is the following: 
\begin{equation}
L_{col} = \chi_{\varepsilon_2}^{d}(d_0, d_1, d_2) \cdot \omega_{\gamma}^{c}(c_0, c_1, c_2) \cdot \rho_{\tau}^{d}(d_0, d_1, d_2),
\end{equation}
where $\chi_{\varepsilon_2}^{d}$ is a depth threshold indicator function with parameter $\varepsilon_2$, $\omega_{\gamma}^{c}$ is a color similarity weight with parameter $\gamma$, and $\rho_{\tau}^{d}$ is a penalty function that evaluates discrepancy for the depth values of collinear 3D points and is the key part of the loss. Here, $c_k$ are the ground truth color vectors and $d_k$ are the rendered depth values of pixels $q_k$ for $k \in \set{0, 1, 2}$.

We denote $u_0,u_1,u_2$ as the unit vectors that correspond to directions of rays for segment pixels in 3D space. If the three 3D points $d_0 u_0$, $d_1 u_1$ and $d_2 u_2$ are expected to lie on single line, assuming $d_1$ as an unknown, one can express the expected value for $d_1$ as:
\begin{equation}
\hat{d_1}= \frac{d_0 d_2 |u_0 \times u_2|}{d_0 |u_0 \times u_1| + d_2 |u_1 \times u_2|}.
\end{equation}

We denote $\Delta d = d_1 - \hat{d}_1$ as the difference between the rendered depth $d_1$ for the midpoint $q_1$ and the expected depth $\hat{d}_1$ based on collinearity of $q_0,q_1,q_2$.
Difference $\Delta d$ is transformed via a hyperbolic tangent to obtain the depth discrepancy loss component: $\rho_{\tau}^d = \tanh{\tau \lvert\Delta d\rvert}$ with $\tau$ controlling the function slope.

The other components improve the quality of the optimization. The indicator $\chi_{\varepsilon_2}^{d}$ is an indicator which verifies that 
$\Delta d$ is not too large relative to the rendered depth values using a hard threshold $\varepsilon$. Particularly, 
\begin{equation}
\chi_{\varepsilon_2}^{d}(d_0, d_1, d_2)=\mathbf{1}\left[\lvert\Delta d\rvert \leq \varepsilon_2 \cdot \min\set{d_0,d_1,d_2}\right].
\end{equation}

Lastly, $\omega_{\gamma}^{c}$ reduces the influence of the loss on less monochromatic regions that were not captured by the edge detector, which we define as: 
\begin{equation}
    \omega_{\gamma}^{c}(c_0, c_1, c_2)=e^{-\frac{\norm{c_1 - c_0}^2 + \norm{c_2 - c_1}^2}{2\gamma^2}}.
\end{equation}

Sampling triplets of pixels is the minimal possible amount of pixels necessary to introduce a notion of local flatness (or any second-order approximation for surfaces). 
Thus our sampling strategy introduces a minimal possible number of co-dependent samples while applying the flattening regularization for ISM.

\subsection{Single-view Depth Denoising (SDD) in NeRF Point Cloud Extraction}
\label{sec:method_depth}

\begin{figure}[tb]
  \begin{subfigure}[t]{.48\linewidth}
    \includegraphics[width=\linewidth]{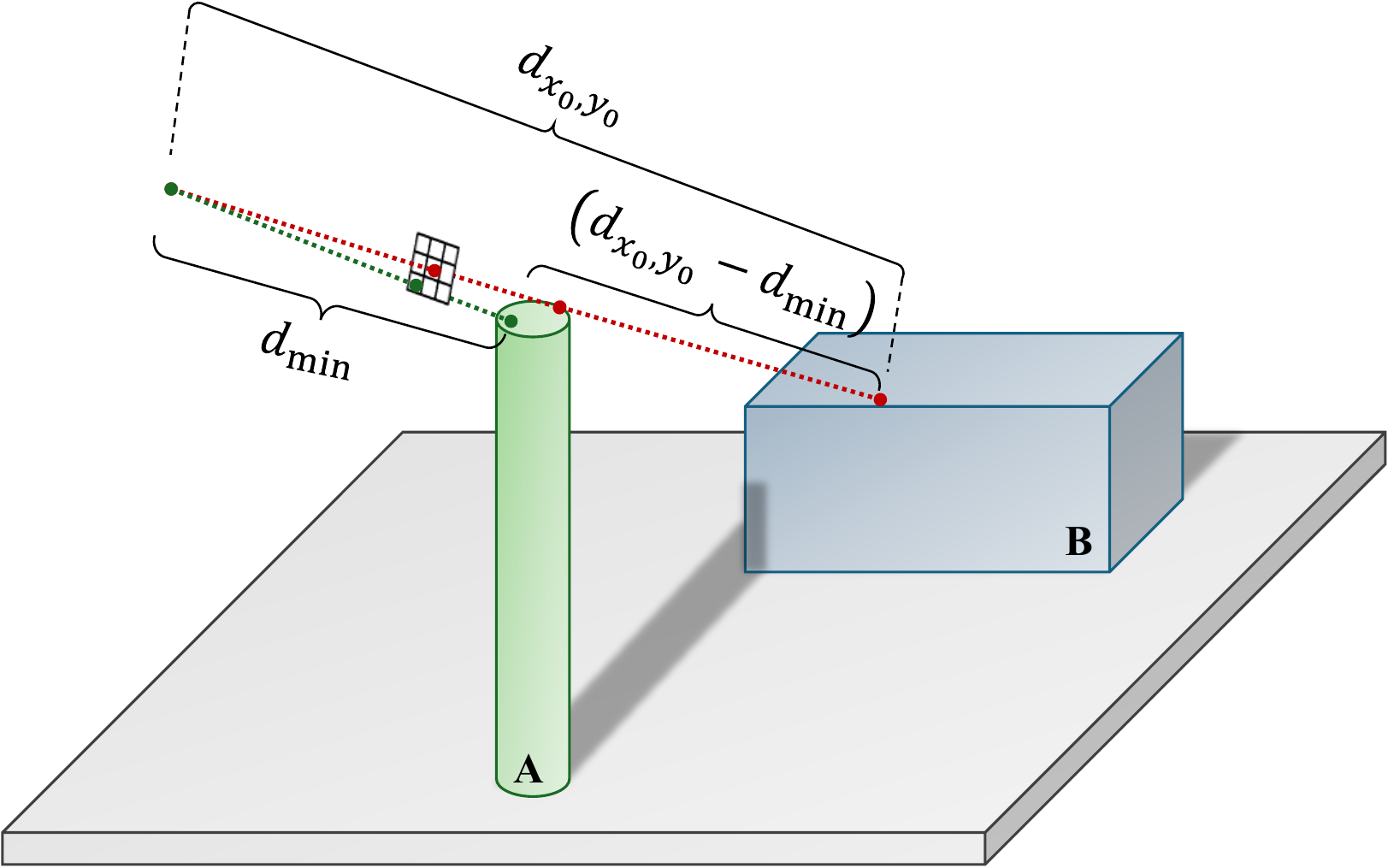}
    \subcaption{Two key depth values in the image patch are compared against one another. 
        The sample is discarded if the tail $(d_{x_0,y_0}-d_{\min})$ exceeds $d_{x_0,y_0}$.
        The 3x3 grid represents the patch of pixels $\Pi$ of the image.}
    \label{fig:depth_problem_unified}
  \end{subfigure}%
  \hfill
  \begin{subfigure}[t]{.48\linewidth}
    \includegraphics[width=\linewidth]{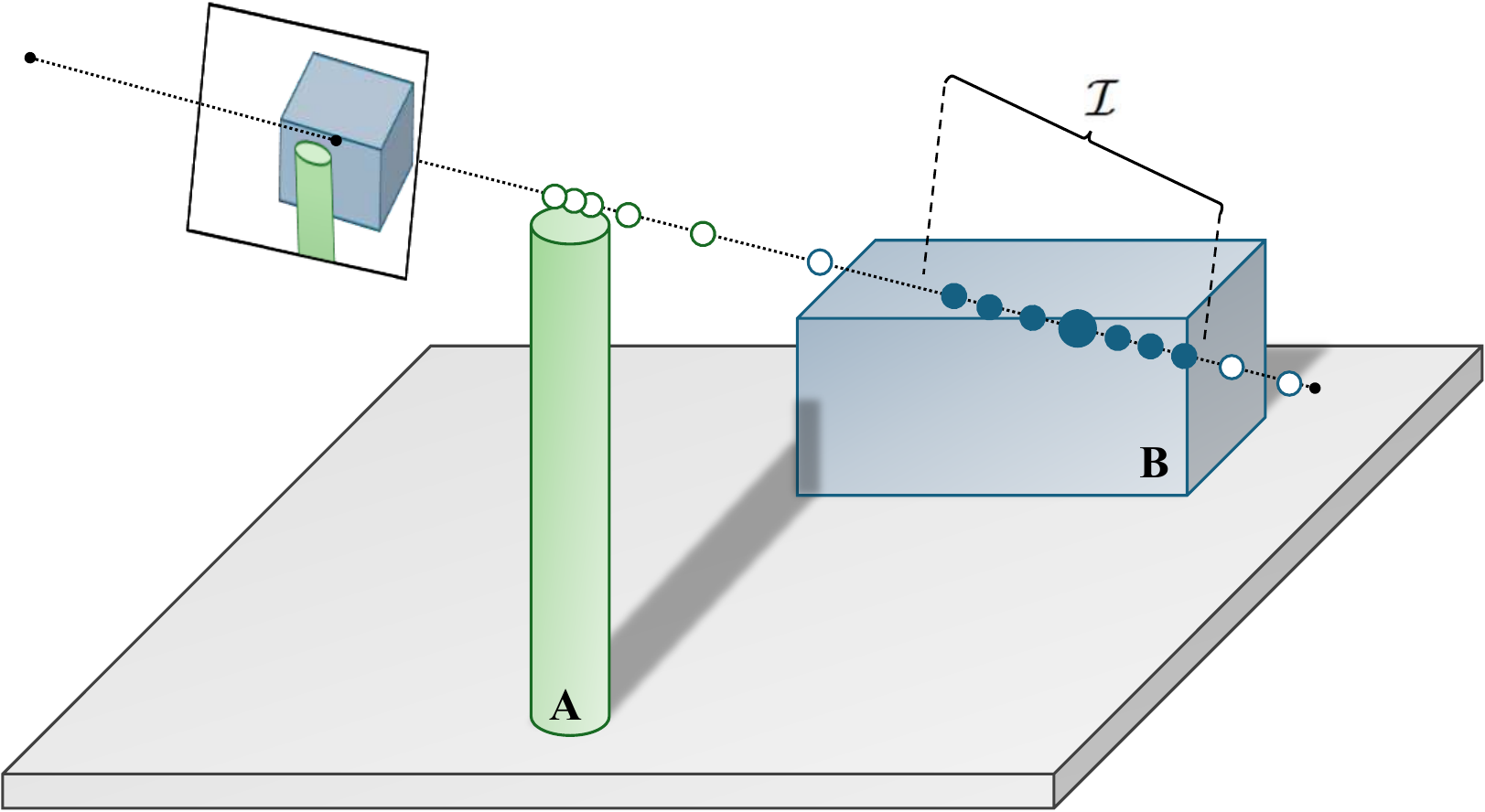}
    \subcaption{Large circle denotes where the depth $d$ was sampled. 
        Smaller filled circles denote the other points with indices in $\mathcal{I}$. 
        Hollow circles denote points with indices outside $\mathcal{I}$ which do not affect the final color computation.}
    \label{fig:color_sharpening}
  \end{subfigure}
  \caption{Example of a single ray passing through objects A and B. Single-view depth denoising is illustrated in (a), color sharpening and denoising is illustrated in (b).}
  \label{fig:depth_color_problems}
\end{figure}

Even with a well-trained NeRF model that would produce realistic renders of the scene, the geometry of point clouds obtained by the current techniques is often noisy: 
sometimes, 3D points that should be part of an object are displaced in an empty volume or on the surface of another object. 
This noise frequently appears where a depth discontinuity would appear in an RGB-D image; 
for example, near the borders of an object that is close to the camera. 
When a ray is cast in a way that passes close to the border of object A, the resulting depth may be estimated to any location in-between the object A and an object B that is located behind, as shown in Figure~\ref{fig:depth_problem_unified} (central ray).



We propose a single-view depth denoising technique, SDD, that helps to significantly reduce the amount of sampled geometric noise. 
Following the standard approach as in \cite{tancik2023nerfstudio}, we construct a point cloud $P$ iteratively by sampling 3D points from a set of images $\mathcal{J}$ with known camera poses. 
Starting from $P \leftarrow \varnothing$, we repeatedly select a random image from $\mathcal{J}$ and uniformly sample pixel coordinates $(x_0,y_0)$ such that a $w \times h$ patch $\Pi$ centered at $(x_0,y_0)$ lies fully inside the image. 
This choice determines the ray direction $u_{x_0,y_0}$ from the camera center $o$ through the pixel. 
$\text{For all}\,(x,y) \in \Pi$, we query the NeRF model to obtain rendered color vectors $c_{x,y}$ and depths $d_{x,y}$, and we keep only the central color $c_{x_0,y_0}$.
To denoise the estimates, the central depth $d_{x_0,y_0}$ is compared to the patch minimum depth $d_{\min}=\min_{(x,y) \in \Pi}d_{x,y}$, as shown in Figure~\ref{fig:depth_problem_unified}. We accept the sample if

\begin{equation}
    (1-\varepsilon_3)\,d_{x_0,y_0} \le d_{\min}
\end{equation}

given a small constant $\varepsilon_3$, ensuring the central ray does not report a depth significantly larger than any of the nearby rays. 
If accepted, we add the corresponding 3D point $o+u_{x_0,y_0}\cdot d_{x_0,y_0}$ with color $c_{x_0,y_0}$ to $P$.
We repeat the sampling procedure until size of the point cloud P reaches a pre-determined value, at which point the final point cloud is created. 


We perform patch sampling in SDD as a two-step process, to decrease the computational overhead.
In the first step we sample and render a batch of rays corresponding to patch centers $(x_0, y_0)$. 
As there are many ways to reject a single sample (such as the final 3D point falling outside a define bounding box), some points get naturally discarded.
Only those rays that would otherwise be picked for a point cloud proceed to the second step, where the rest of the rays from the patches are sampled. 

\subsection{Color Sharpening and Denoising (CSD) in NeRF Point Cloud Extraction}
\label{sec:method_color}

NeRF standard volume rendering aggregates radiance along the entire ray. That is appropriate for image synthesis, but becomes problematic for point cloud extraction, where a color of a point should be view-independent.
View-dependent effects such as color “bleeding” (\eg, a white surface observed through blue glass) can bias sampled point colors away from their true surface appearance.


This section covers the color sharpening and denoising modification for point cloud extraction, CSD, that we perform to obtain the $c_{x_0,y_0}$ color vector which appeared in Section~\ref{sec:method_depth}.

The ray passing through the pixel $(x_0, y_0)$ is discretized into points $\set{p_i}_{i=1}^n$. 
The NeRF model is queried at those points to obtain colors $\set{c_i}_{i=1}^n$ and densities $\set{\sigma_i}_{i=1}^n$.
Using the volume-rendering procedure, the densities are converted to non-negative weights $\set{w_i}_{i=1}^n$ with $w_i\in[0,1]$ and $\sum_i w_i \le 1$. 
Depth is rendered as $d = \|o-p_{i_0}\|$, where $i_0$ is the first index where the cumulative weight sum exceeds 0.5; if no such $i_0$ exists, we set $d=\infty$ and reject the sample.

In most NeRF methods, the standard volume rendering approach to render color is: $c=\sum_{i=1}^n w_i\,c_i$. 
For rendering with CSD, we set $d_i=\norm{o-p_i}$ and select indices of points whose depths lie near the rendered depth: $\mathcal{I}=\{i \mid d(1-\varepsilon_4)\le d_i \le d(1+\varepsilon_4)\}$ for a threshold constant $\varepsilon_4$. If $\mathcal{I}=\varnothing$, we reject the sample. 
We renormalize weights within this segment as 
\begin{equation}
    \tilde{w}_i = \frac{\sum_{j=1}^{n} w_j}{\sum_{j\in \mathcal{I}} w_j}w_i, \quad\forall i\in \mathcal{I}
\end{equation}
and compute the color of the point as $c=\sum_{i\in \mathcal{I}}\tilde{w}_i\,c_i$. 
Figure~\ref{fig:color_sharpening} illustrates the sample contributions to the color computation.


\section{Experiments}
\label{sec:experiments}

To evaluate the metric accuracy of NeRF-sampled 3D point clouds, we performed experiments in real-world environments. 
We generate reference data for our evaluation by performing scans with a BLK360 LiDAR scanner~\cite{leicaBLK360}. 
Point clouds produced by BLK360 are metrically scaled. All other generated point clouds obtain their metric scale from geo-registration, 
wherein the image poses estimated in SfM are registered to the corresponding UAV~\cite{djiMavic3Pro} images' positions from RTK-GPS metadata, using COLMAP's model aligner~\cite{schonberger2016structure}.

We compare our proposed CLEAR-NeRF against Nerfacto~\cite{tancik2023nerfstudio} (\texttt{v1.1.5}), a state-of-the-art neural radiance field method that serves as our first baseline. 
CLEAR-NeRF extends Nerfacto with the techniques described in Section~\ref{sec:method}. 
Both NeRF-based methods incorporate monocular depth priors estimated using Depth Anything v2~\cite{yang2024depth} for additional depth supervision, following the depth-supervised NeRF training strategy with ranking loss. 
Both methods share common hyperparameters, such as a batch size of 8064 rays and 60000 training iterations. 
All methods utilize GLOMAP (\texttt{v1.0.0})\cite{pan2024glomap} for camera alignment, which are reoptimized during NeRF training. 
We also compare against classical MVS implemented in COLMAP~\cite{schonberger2016pixelwise} in the quantitative evaluation as the second baseline.

Throughout our experiments, we use the following hyperparameters: the maximum number of focus areas is set to $M = 5$, with $n=20$ neighbors used to compute distance to camera with the highest concentraction of rays $t_{\kappa}$. 
For frustum-based distance computation, we use an angular size of $\alpha=10^\circ$. 
The HDBSCAN clustering algorithm is configured with \texttt{min\_cluster\_size} set to 20 and single cluster selection enabled. 
For collinearity loss, we set the slope parameter $\tau=4$ and the color sharpening scale parameter $\gamma=0.1$ for color values normalized to $[0, 1]$. 
For SDD point cloud extraction, we use $3 \times 3$ patches ($w=h=3$).
The various depth thresholds are uniformly set to $\varepsilon_2=\varepsilon_3=\varepsilon_4=0.0025$.

\begin{figure}[tb]
    \centering
    \begin{subfigure}[t]{0.45\textwidth}
        \centering
        \includegraphics[width=\linewidth, height=0.3\textheight, keepaspectratio, trim=0 65pt 0 19pt, clip]{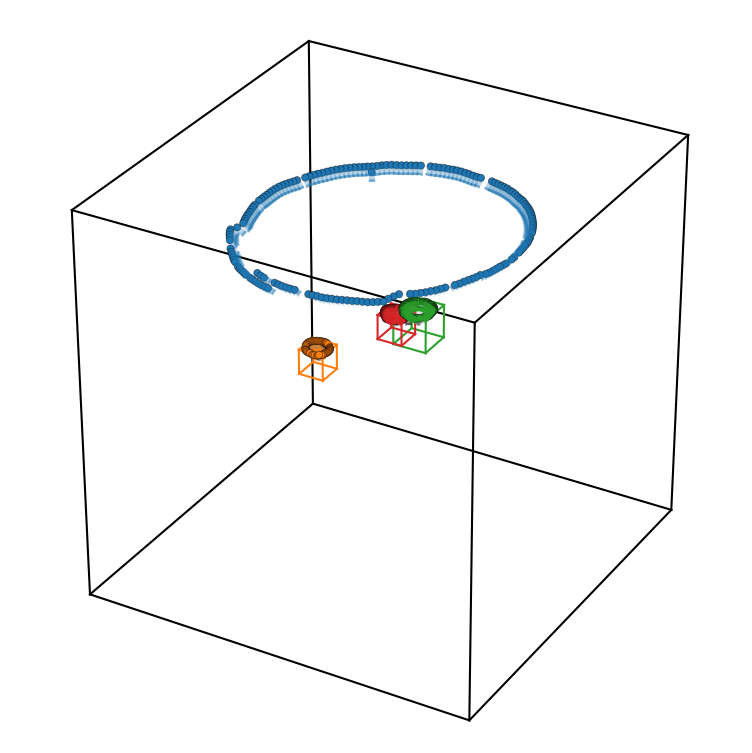}
        \caption{Flight path with 3 focus areas}
        \label{fig:orbits_437}
    \end{subfigure}
    \begin{subfigure}[t]{0.45\textwidth}
        \centering
        \includegraphics[width=\linewidth, height=0.3\textheight, keepaspectratio]{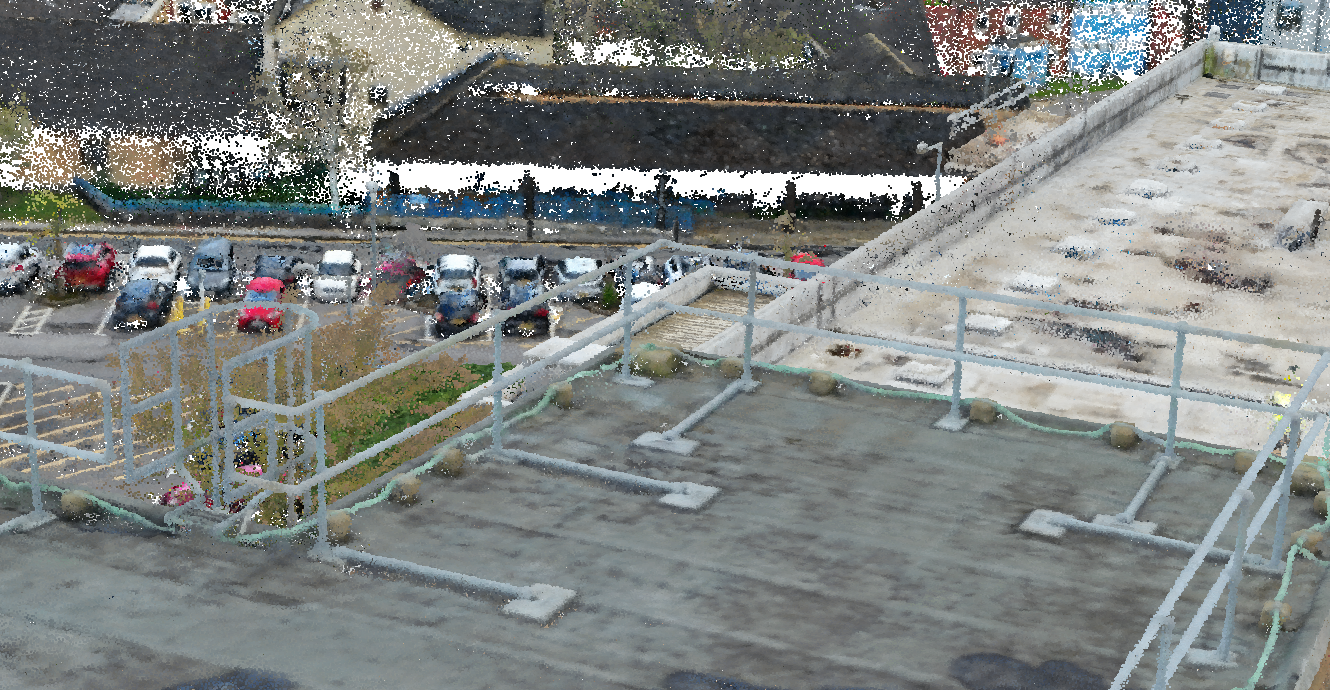}
        \caption{CLEAR-NeRF (all modules)}
        \label{fig:20mil-focus}
    \end{subfigure}

    \begin{subfigure}[t]{0.45\textwidth}
        \centering
        \includegraphics[width=\linewidth, height=0.3\textheight, keepaspectratio]{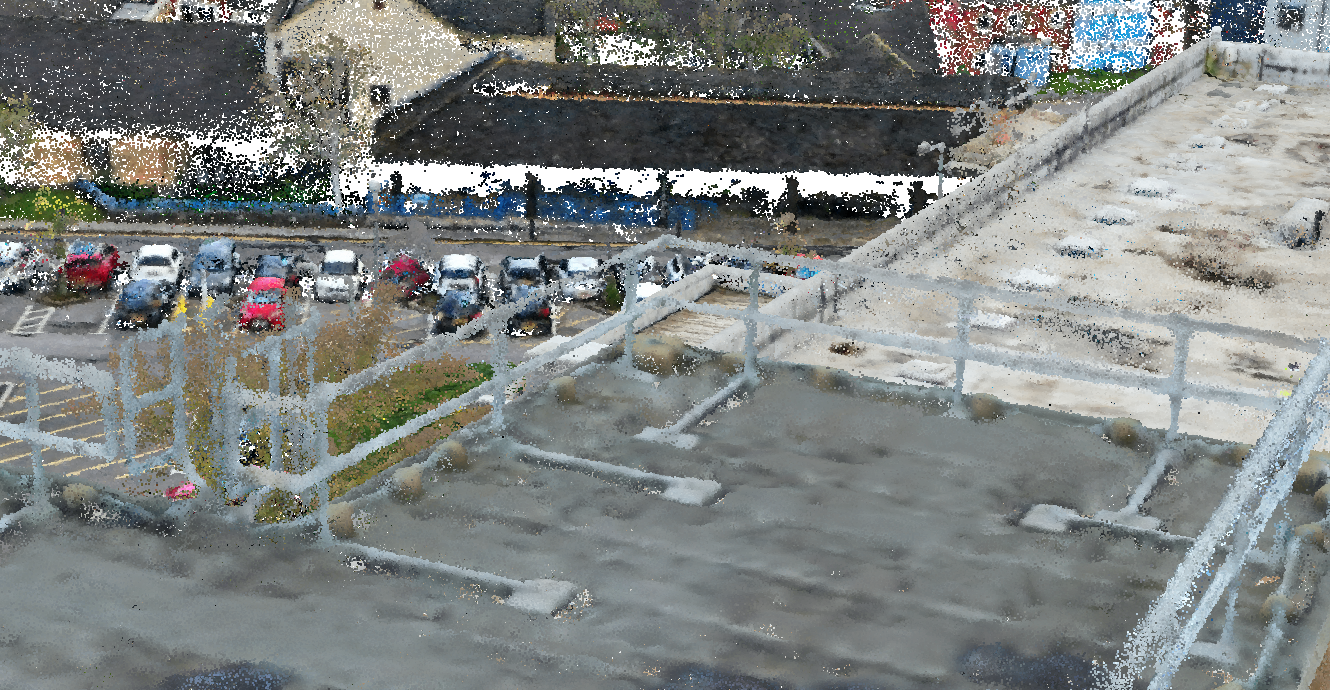}
        \caption{CLEAR-NeRF w/o LRF (\ref{sec:method_focus_areas})}
        \label{fig:20mil-uniform}
    \end{subfigure}
    \begin{subfigure}[t]{0.45\textwidth}
        \centering
        \includegraphics[width=\linewidth, height=0.3\textheight, keepaspectratio]{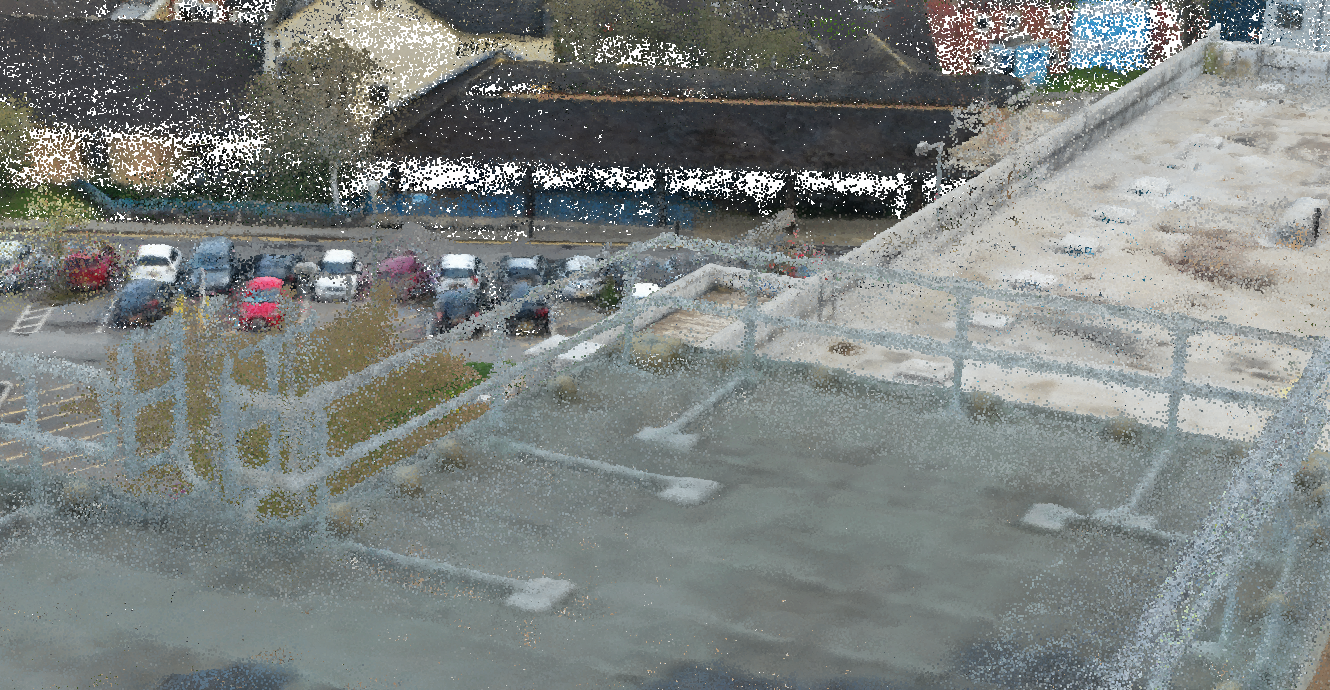}
        \caption{Baseline Nerfacto}
        \label{fig:20mil-baseline}
    \end{subfigure}
    \caption{Visual comparison of sampled 3D point cloud quality for LRF.}
    \label{fig:compare-focus}
\end{figure}

\subsection{Ablation Studies}
In this section we evaluate contribution of the individual components of the proposed CLEAR-NeRF scheme. 
Results from the baseline Nerfacto are also provided for reference.

\textbf{LRF Scene Reconstruction.}
Here we present an outdoor capture performed with a UAV on different levels: one orbit 60-67 meters above the building and several small orbits 6-8 meters above the roof. 
The flight pattern is depicted in Figure~\ref{fig:orbits_437}, with $M_0=3$ relevant high-fidelity areas determined by LRF.
Figure~\ref{fig:20mil-focus} shows the reconstruction obtained using the four modules presented in CLEAR-NeRF, while Figure~\ref{fig:20mil-uniform} shows the result of disabling LRF module.
Clear visual differences can be observed, particularly in the railing and floor bars, which exhibit noticeably higher fidelity when applying LRF.

\begin{figure}[tb]
    \def\sfwidth{0.32} 
    \centering
    \begin{subfigure}[t]{\sfwidth\textwidth}
        \centering
        \includegraphics[width=\linewidth]{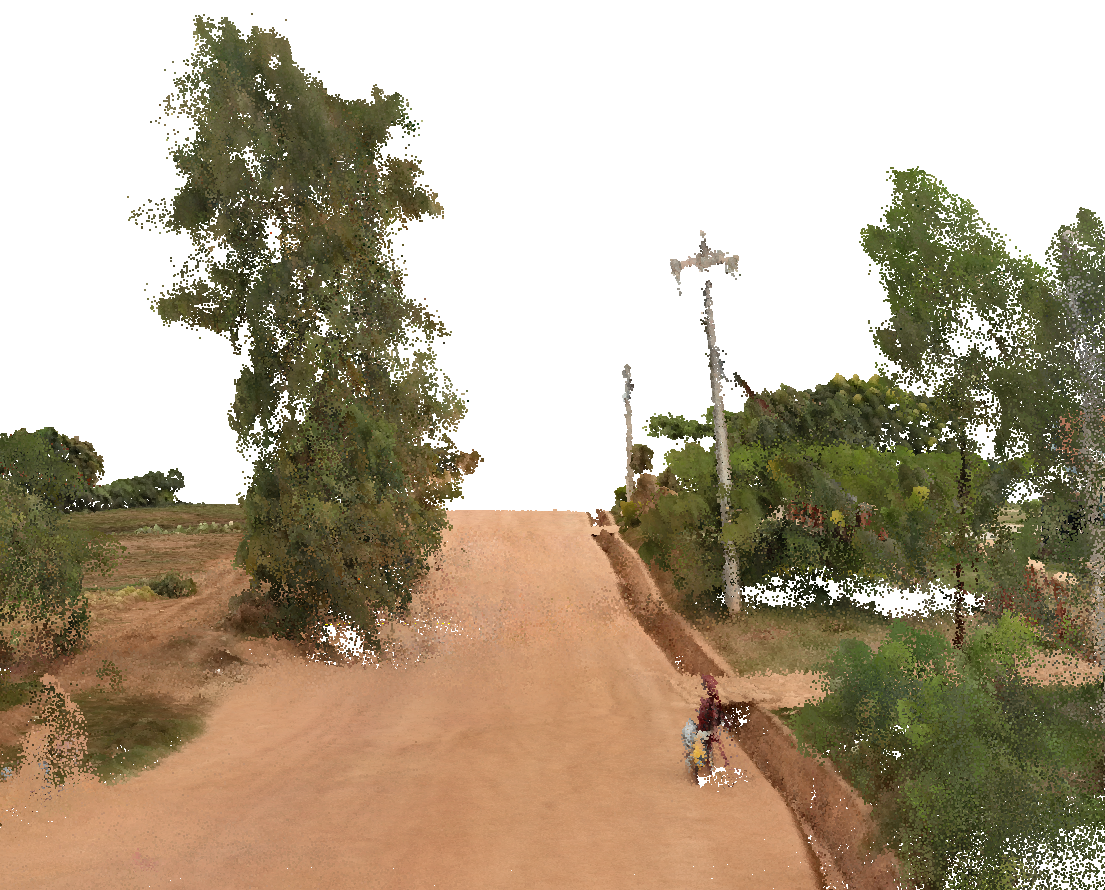}
        \caption{All modules}
        \label{fig:rwanda-1111}
    \end{subfigure}
    \begin{subfigure}[t]{\sfwidth\textwidth}
        \centering
        \includegraphics[width=\linewidth]{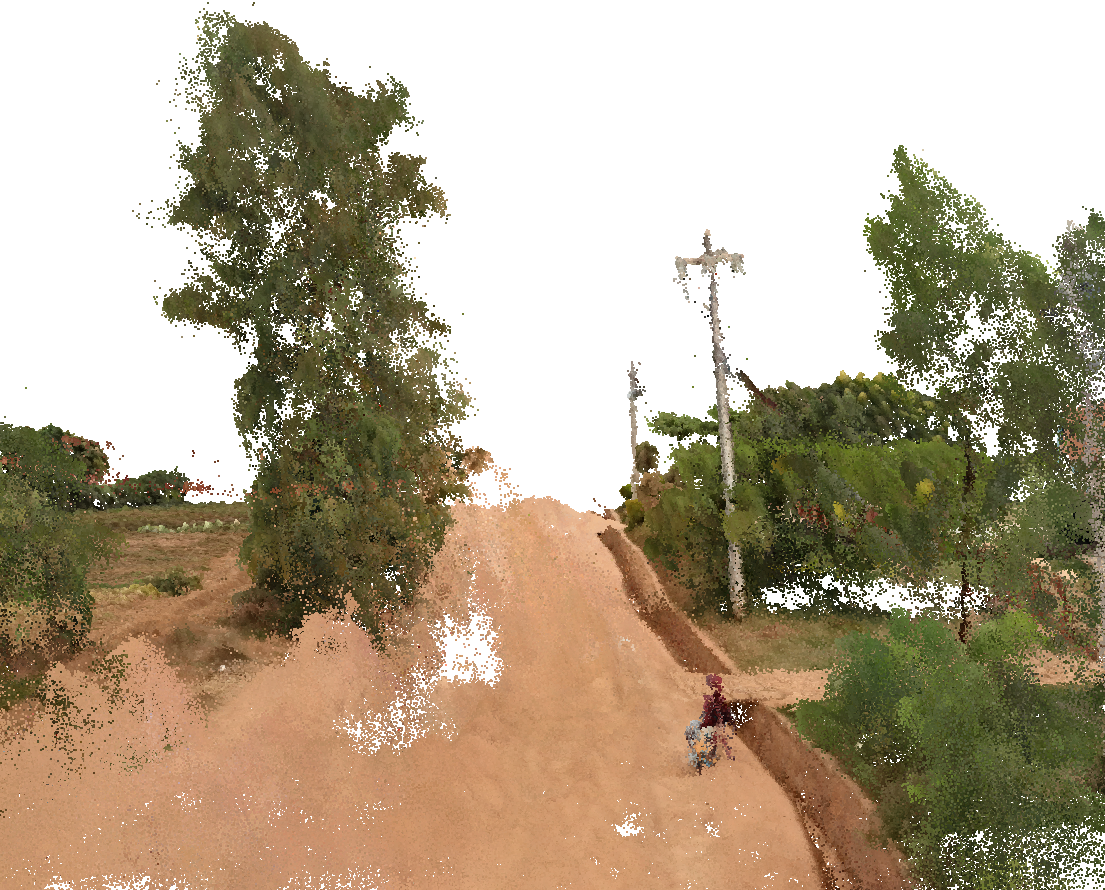}
        \caption{w/o ISM (\ref{sec:method_collinearity})}
        \label{fig:rwanda-1011}
    \end{subfigure}\\
    \begin{subfigure}[t]{\sfwidth\textwidth}
        \centering
        \includegraphics[width=\linewidth]{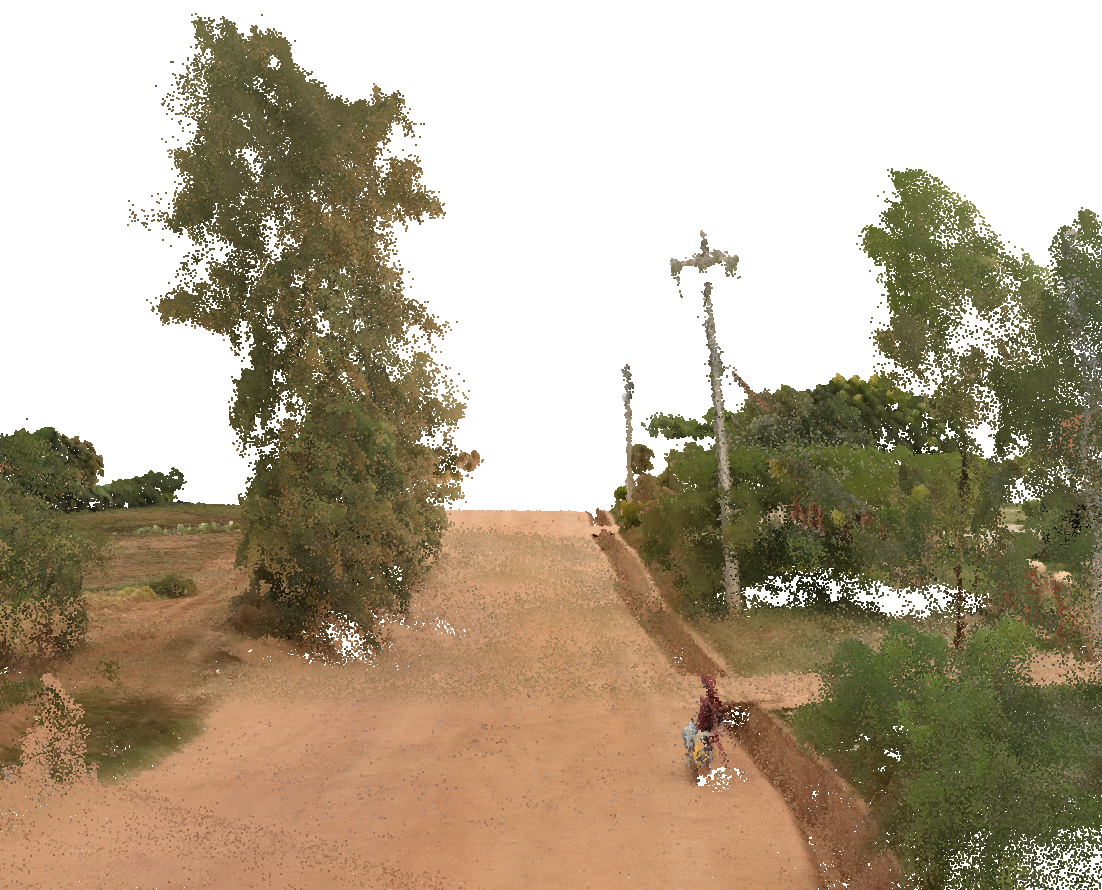}
        \caption{w/o CSD (\ref{sec:method_color})}
        \label{fig:rwanda-1110}
    \end{subfigure}
    \begin{subfigure}[t]{\sfwidth\textwidth}
        \centering
        \includegraphics[width=\linewidth]{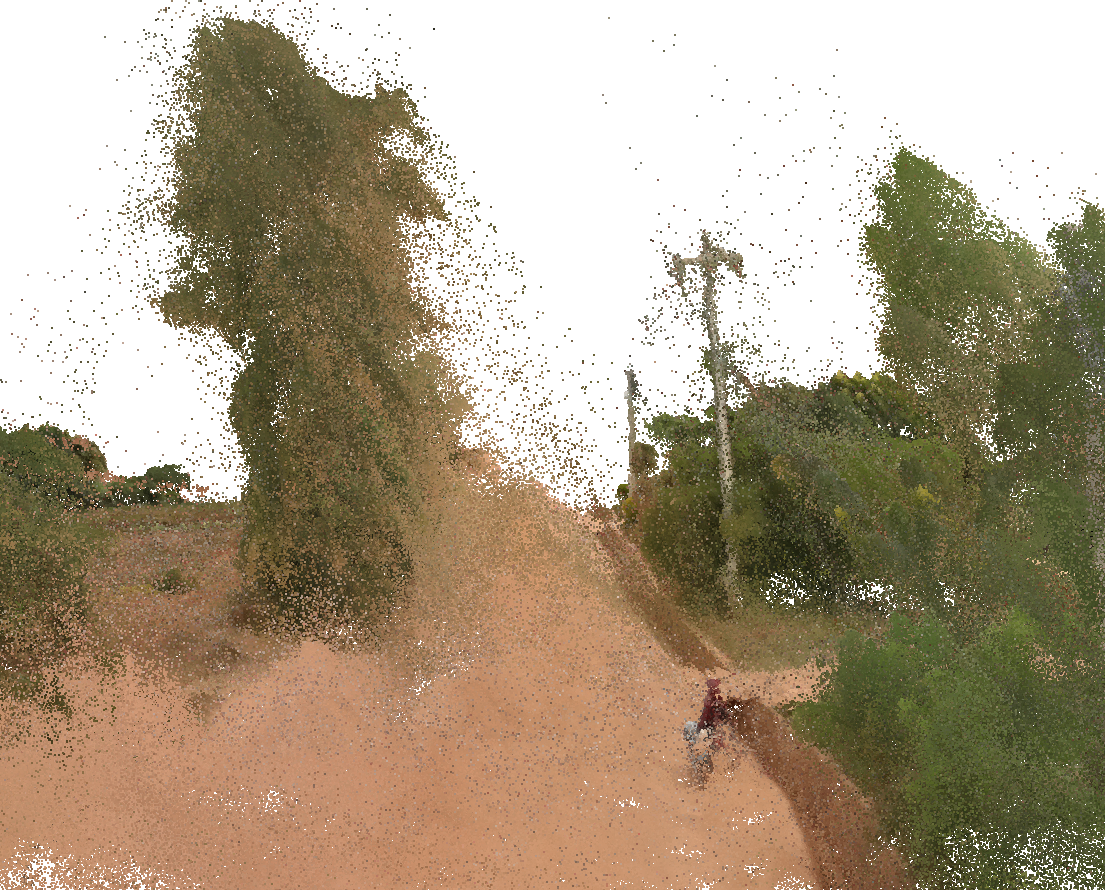}
        \caption{Baseline Nerfacto}
        \label{fig:rwanda-baseline}
    \end{subfigure}
    \caption{Visual comparison of sampled 3D point cloud quality for ISM and CSD.}
    \label{fig:compare-accuracy}
\end{figure}

\textbf{Geometric Accuracy of 3D Reconstruction.}
The drone capture of an urban area presented in Figures~\ref{fig:compare-accuracy} has been performed at various altitudes 35 and 75 meters above the ground level.
The figure shows a visual comparison of the reconstruction performance under different configurations.
In Figure~\ref{fig:rwanda-1111}, the four techniques proposed in CLEAR-NeRF are jointly applied.
Figure~\ref{fig:rwanda-1011} shows the same scene without ISM, where the ground exhibits noticeable deformation. 
Figure \ref{fig:rwanda-1110} illustrates the result of disabling CSD, which leads to visible noise, 
particularly green noise projected from the tree onto the ground. This noise is effectively suppressed when CSD is employed.

\begin{figure}[tb]
    \def\figheight{125pt}
    \centering
    \begin{subfigure}[t]{0.31\textwidth}
        \centering
        \includegraphics[width=\linewidth, height=\figheight, keepaspectratio, trim=0 30pt 0 0pt, clip]{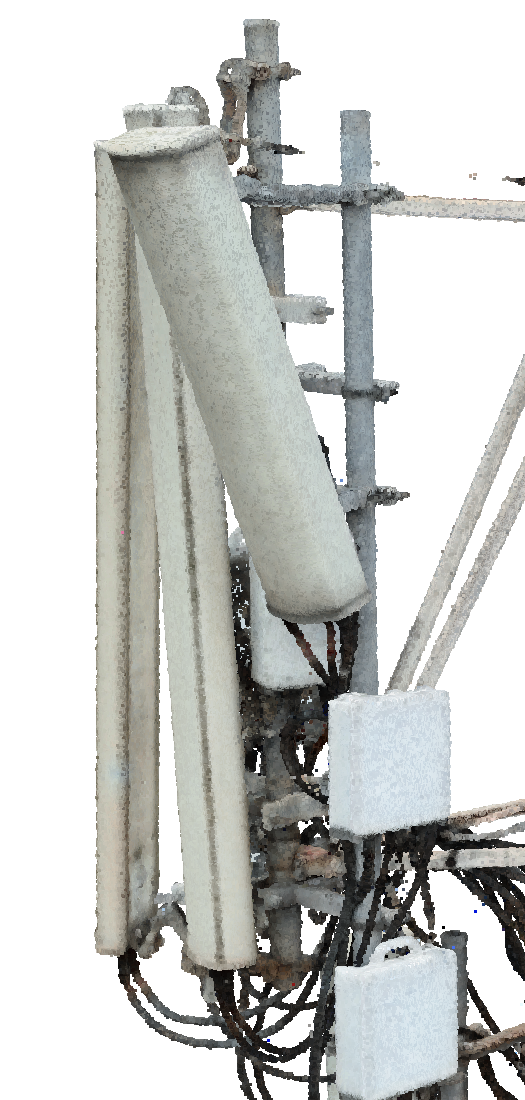}
        \caption{All modules}
        \label{fig:antenna-1111}
    \end{subfigure}
    \begin{subfigure}[t]{0.31\textwidth}
        \centering
        \includegraphics[width=\linewidth, height=\figheight, keepaspectratio, trim=0 30pt 0 0pt, clip]{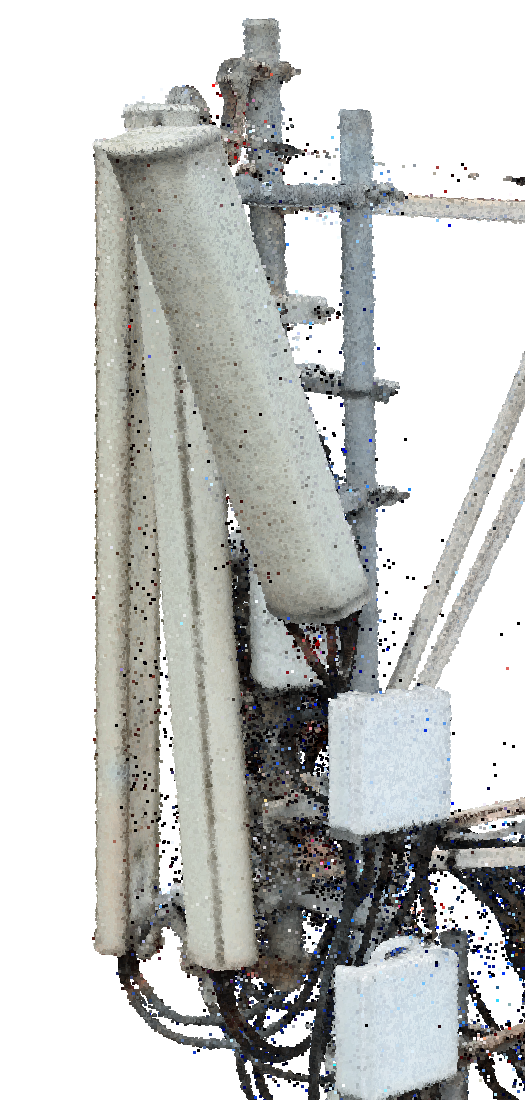}
        \caption{w/o SDD (\ref{sec:method_depth})}
        \label{fig:antenna-1101}
    \end{subfigure}
    \begin{subfigure}[t]{0.31\textwidth}
        \centering
        \includegraphics[width=\linewidth, height=\figheight, keepaspectratio, trim=0 30pt 0 0pt, clip]{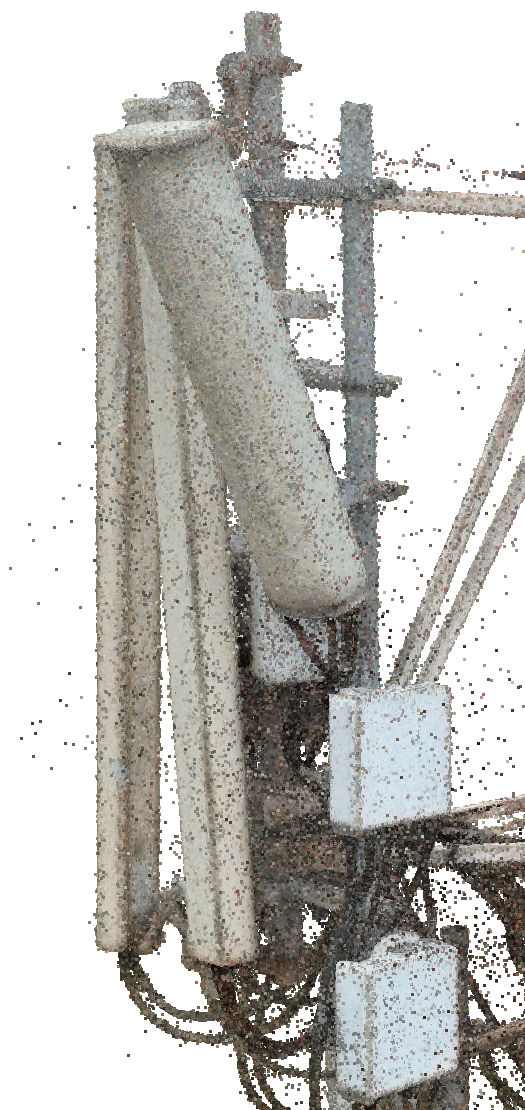}
        \caption{\centering Baseline Nerfacto}
        \label{fig:antenna-0000}
    \end{subfigure}
    \caption{Visual comparison of sampled 3D point cloud quality for SDD.}
    \label{fig:compare-depth}
\end{figure}

Figure~\ref{fig:compare-depth} displays a close-up (captured at 5 meters proximity with a UAV) of an object with multiple overlapping surfaces to highlight how noise impacts such scenes. 
Figure~\ref{fig:antenna-1111} presents the result using CLEAR-NeRF, 
which demonstrates a significant reduction in noise compared to the case without SDD, 
shown in Figure~\ref{fig:antenna-1101}. Finally, Figure~\ref{fig:antenna-0000} represents the result of the reconstruction with our baseline model Nerfacto.

\subsection{Comparison to LiDAR Point Clouds}

\begin{figure}[htb]
    \centering
    \begin{subfigure}[t]{0.24\textwidth}
        \centering
        \includegraphics[width=\linewidth]{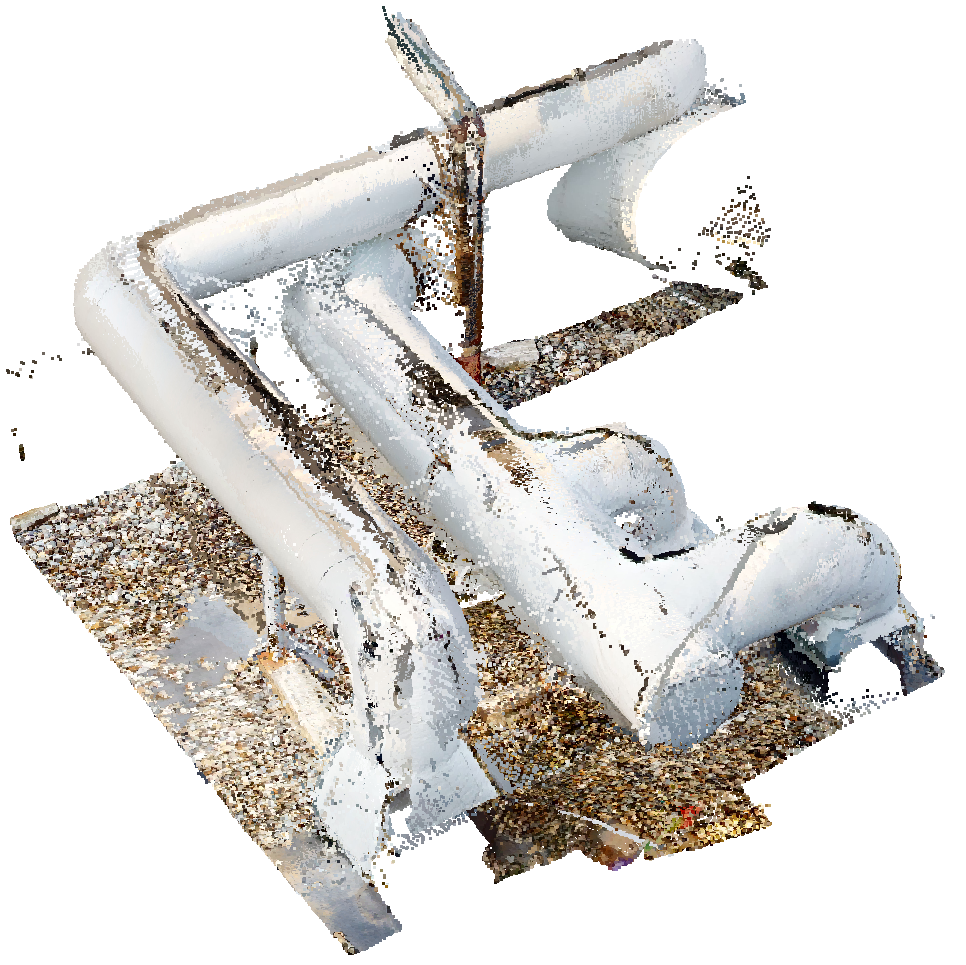}
        \includegraphics[width=\linewidth]{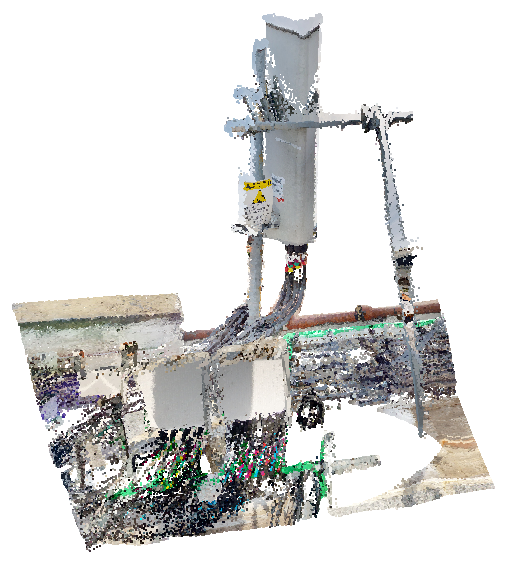}
        \caption{LiDAR reference}
        \label{fig:pipe-lidar}
    \end{subfigure}
    \begin{subfigure}[t]{0.24\textwidth}
        \centering
        \includegraphics[width=\linewidth]{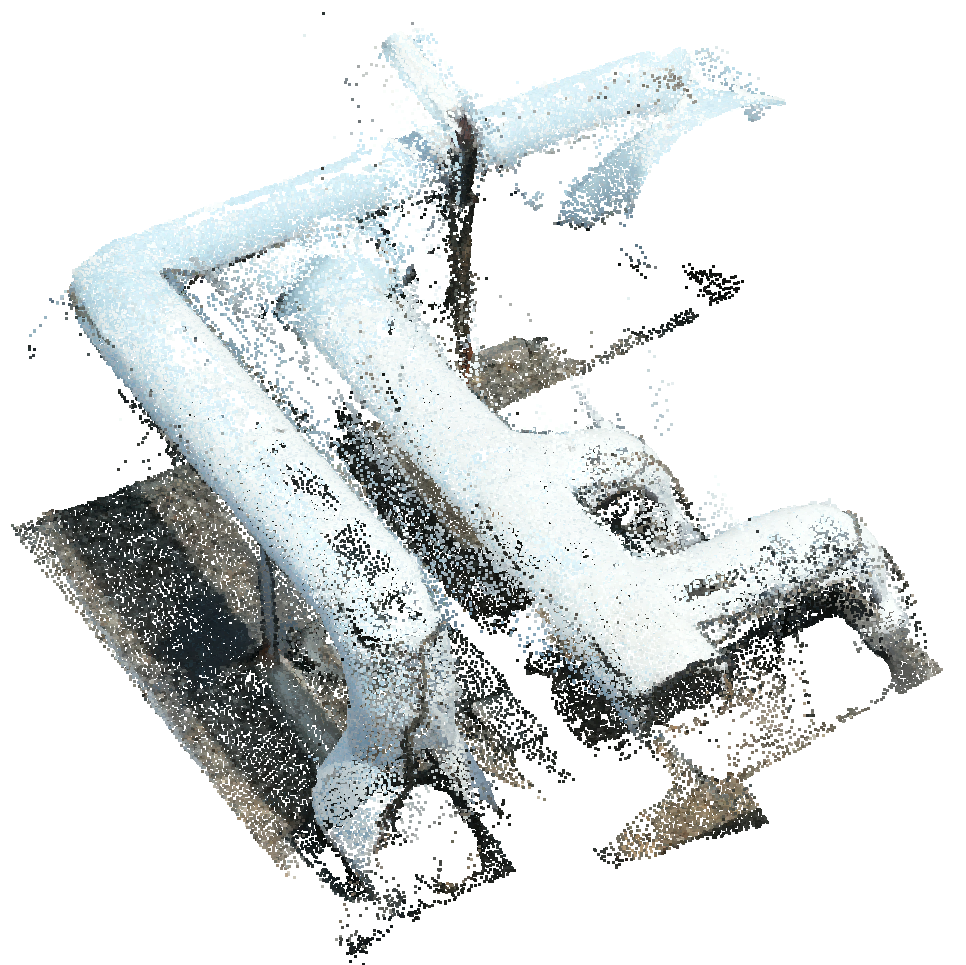}
        \includegraphics[width=\linewidth]{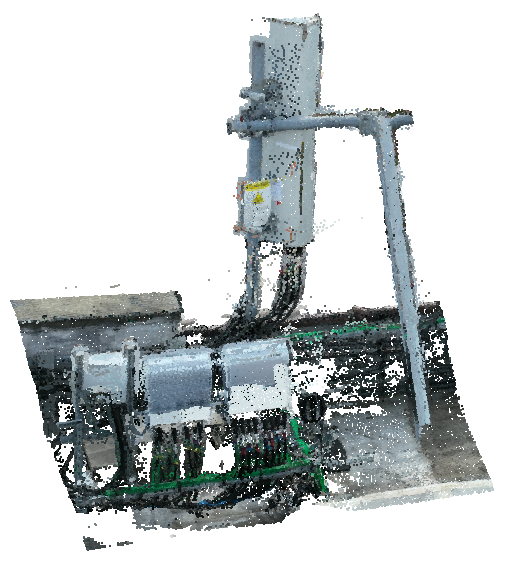}
        \caption{MVS}
        \label{fig:pipe-fused}
    \end{subfigure}
    \begin{subfigure}[t]{0.24\textwidth}
        \centering
        \includegraphics[width=\linewidth]{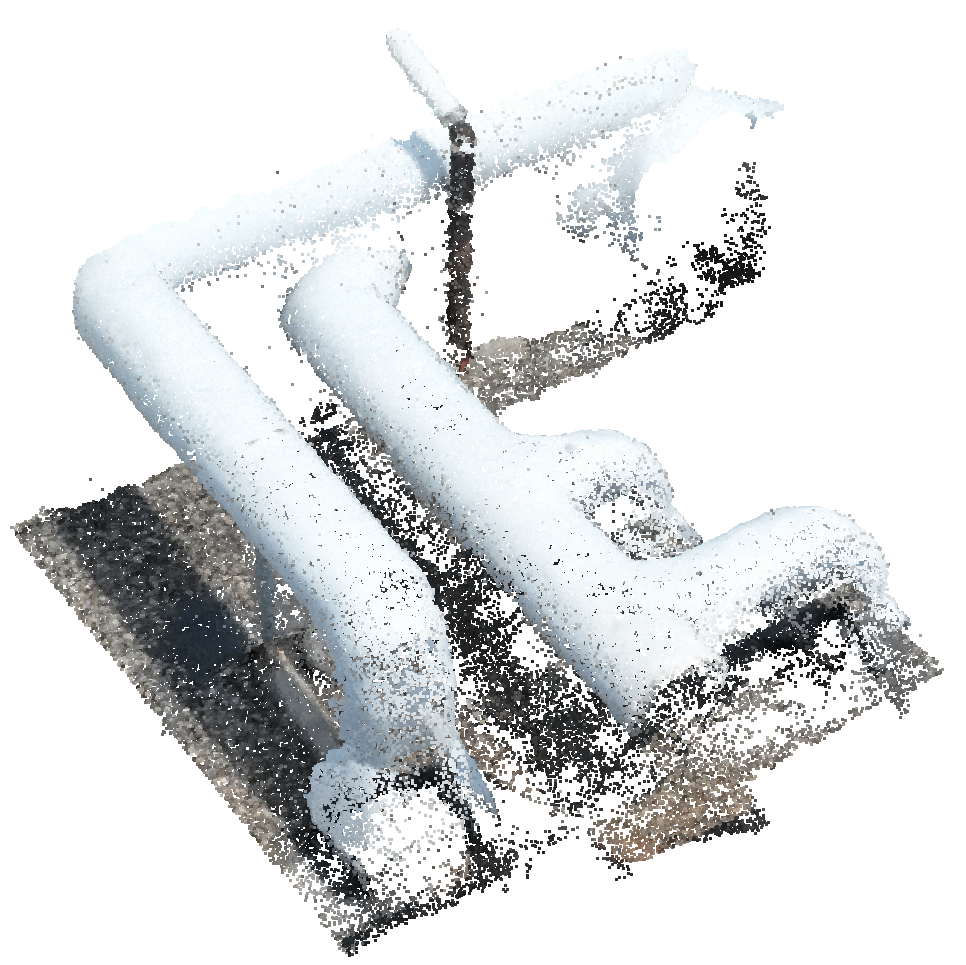}
        \includegraphics[width=\linewidth]{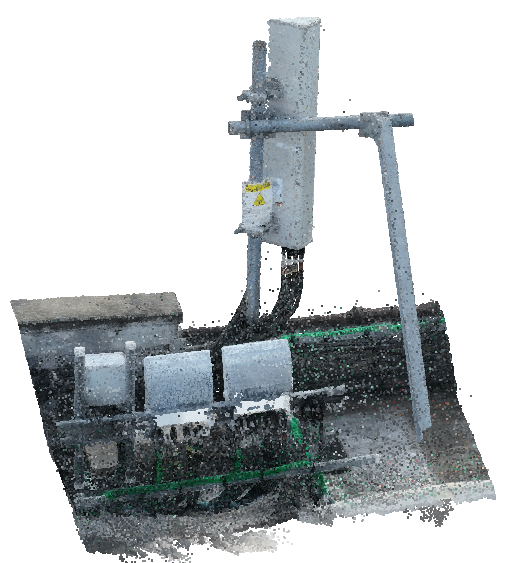}
        \caption{Baseline Nerfacto}
        \label{fig:pipe-0000}
    \end{subfigure}
    \begin{subfigure}[t]{0.24\textwidth}
        \centering
        \includegraphics[width=\linewidth]{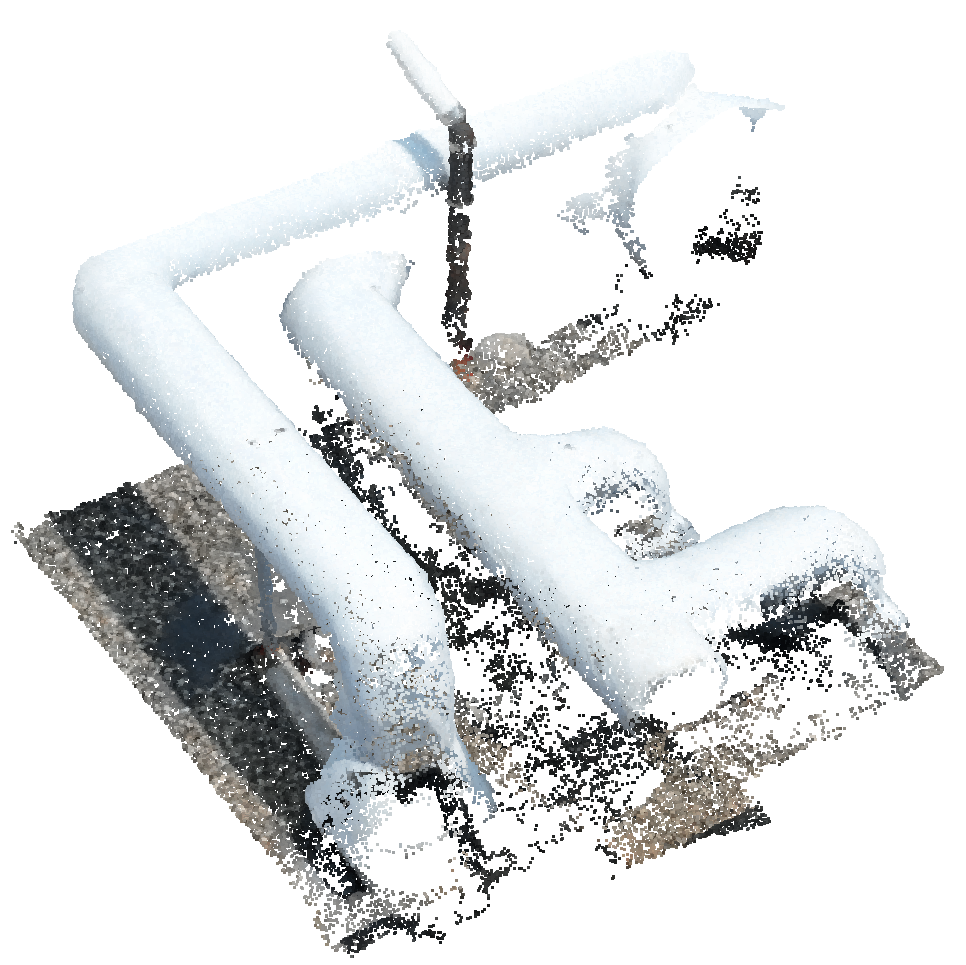}
        \includegraphics[width=\linewidth]{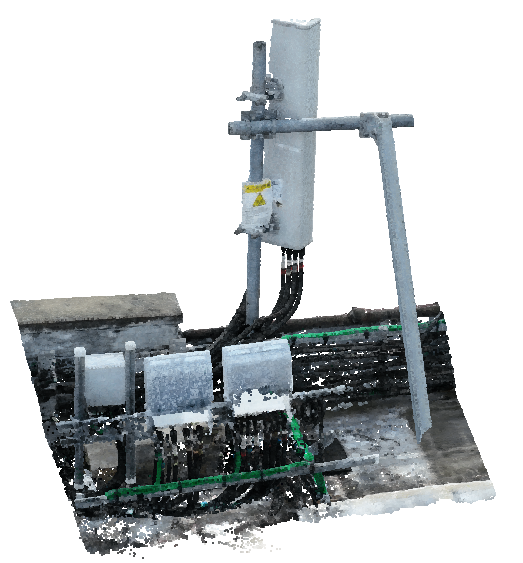}
        \caption{CLEAR-NeRF}
        \label{fig:pipe-1111}
    \end{subfigure}
    \caption{Comparison of point cloud reconstruction methods against LiDAR reference on two selected objects.}
    \label{fig:pipe-comparison}
\end{figure}

To quantitatively evaluate the 3D point cloud reconstruction quality, we compare the generated point clouds against high-precision LiDAR scans in real-world scenarios.
Objects of interest, as illustrated in Figure~\ref{fig:pipe-comparison}, are extracted from the scene, and closeness of LiDAR and UAV image-based point clouds is evaluated.

Table~\ref{tab:lidar-eval} reports quantitative metrics including Chamfer distance, Hausdorff distance, and F-Score, between the LiDAR point clouds and the tested methods.
CLEAR-NeRF achieves the best F-Score in 6 out of 7 cases, demonstrating superior overall geometric accuracy compared to both classical MVS and the baseline Nerfacto method.

\begin{table}[tb]
\centering
\caption{Reconstructed 3D point cloud comparison to LiDAR reference.}
\label{tab:lidar-eval}
\begin{tabular}{c|l|p{60pt}p{60pt}p{60pt}} 
\toprule
 & Metrics    & ~MVS   & Nerfacto & CLEAR-NeRF \\ \midrule
\multirow{3}{*}{\#1}  & Chamfer \hfill$\searrow~$   & ~0.122 & \textbf{0.105}    & \underline{0.106}      \\ 
                             & Hausdorff \hfill$\searrow~$ & ~0.577 & \textbf{0.407}    & \underline{0.619}      \\ 
                             & F-Score \hfill$\nearrow~$   & ~80.577 & \underline{84.254}  & \textbf{84.832}     \\ \midrule
\multirow{3}{*}{\#2}  & Chamfer \hfill$\searrow~$   & ~\textbf{0.061} & 0.078    & \underline{0.067}      \\ 
                             & Hausdorff \hfill$\searrow~$ & ~\underline{0.557} & 8.867    & \textbf{0.471}      \\ 
                             & F-Score \hfill$\nearrow~$   & ~\underline{94.818} & 93.282  & \textbf{95.365}     \\ \midrule
\multirow{3}{*}{\#3}  & Chamfer \hfill$\searrow~$   & ~0.181 & \textbf{0.161}    & \underline{0.165}      \\ 
                             & Hausdorff \hfill$\searrow~$ & ~\textbf{0.947} & \underline{0.960}    & 0.962      \\ 
                             & F-Score \hfill$\nearrow~$   & ~78.548 & \underline{80.677}  & \textbf{81.223}     \\ \midrule
\multirow{3}{*}{\#4}  & Chamfer \hfill$\searrow~$   & ~\textbf{0.071} & 0.082    & \underline{0.074}      \\ 
                             & Hausdorff \hfill$\searrow~$ & ~\underline{0.636} & 3.524    & \textbf{0.624}      \\ 
                             & F-Score \hfill$\nearrow~$   & ~\underline{91.870} & 91.790  & \textbf{92.617}     \\ \midrule
\multirow{3}{*}{\#5}  & Chamfer \hfill$\searrow~$   & ~\underline{0.182} & 0.194    & \textbf{0.163}      \\ 
                             & Hausdorff \hfill$\searrow~$ & ~\underline{1.250} & 1.252    & \textbf{1.148}      \\ 
                             & F-Score \hfill$\nearrow~$   & ~\underline{74.538} & 71.806  & \textbf{75.637}     \\ \midrule
\multirow{3}{*}{\#6}  & Chamfer \hfill$\searrow~$   & ~0.025 & \textbf{0.020}    & \textbf{0.020}      \\ 
                             & Hausdorff \hfill$\searrow~$ & ~\underline{0.189} & 0.215    & \textbf{0.161}      \\ 
                             & F-Score \hfill$\nearrow~$   & ~\textbf{99.972} & 99.823  & \underline{99.938}     \\ \midrule
\multirow{3}{*}{\#7}  & Chamfer \hfill$\searrow~$   & ~0.059 & \underline{0.049}    & \textbf{0.047}      \\ 
                             & Hausdorff \hfill$\searrow~$ & ~0.311 & \underline{0.307}    & \textbf{0.304}      \\ 
                             & F-Score \hfill$\nearrow~$   & ~94.530 & \underline{95.544}  & \textbf{95.858}     \\ \bottomrule
\end{tabular}
\end{table}

\subsection{Runtime Analysis} 

We evaluate the computational overhead of CLEAR-NeRF compared to the baseline Nerfacto on an NVIDIA A100 GPU.

LRF focus area detection introduces negligible computational cost, while each additional focus area adds approximately 30\% to the base training and export time.
ISM incurs minimal overhead, limited to the edge detection preprocessing step. CSD does not introduce any additional processing cost.

The two-step sampling procedure for SDD (Section~\ref{sec:method_depth}) ensures that computational cost scales with the final point cloud size.
This eliminates unnecessary computation on rejected samples (\eg, rays falling outside scene bounds or at infinity), with benefits particularly significant when many image pixels correspond to background regions.
For the dataset in Figure~\ref{fig:compare-depth}, exporting 20 million points takes 300 seconds without SDD and 780 seconds with SDD using $3\times3$ patches (a 480 second overhead fixed to the point cloud size).
In contrast, a naive implementation of SDD requires approximately 2700 seconds for the same export.

\section{Conclusion}
\label{sec:conclusion}

In this paper, we introduce CLEAR-NeRF, a novel method allowing for accurate 3D reconstruction in unbounded, complex scenes with imperfect lighting conditions. 
The core components of the proposed architecture are \textit{a}) automated selection and training of multi-regions of interest, 
\textit{b}) novel collinearity constraints for surface smoothness, \textit{c}) improved 3D point cloud sampling with depth and color denoising.

We conduct experiments to investigate accuracy of the proposed method in real-world environment by performing 3D reconstruction on 
UAV image sets from urban scenes. We use LiDAR scans to establish ground truth for benchmarking the proposed improvements in the 
baseline NeRF architecture. Initial experiments indicate superior performance of CLEAR-NeRF over legacy SfM-MVS and baseline NeRF.


\bibliographystyle{splncs04}
\bibliography{main}
\end{document}